\documentclass{article}
\usepackage[nonatbib, preprint]{neurips_2026}

\usepackage[utf8]{inputenc} 
\usepackage[T1]{fontenc}    
\usepackage{hyperref}       
\usepackage{url}            
\usepackage{booktabs}       
\usepackage{amsfonts}       
\usepackage{nicefrac}       
\usepackage{microtype}      
\usepackage{xcolor}         

\usepackage{enumitem}
\usepackage{comment}

\usepackage{todonotes}
\title{Universal Approximation Theorem for Metaplectic Neural Operators}

\author{%
  Ahmed Abdeljawad \\
  RICAM, Austrian Academy of Sciences \\
  Altenberger Straße 69, A-4040 Linz, Austria\\
  \texttt{Ahmed.Abdeljawad@oeaw.ac.at}
  \And
  Malek Yacine Rezoug \\
  RICAM, Austrian Academy of Sciences \\
  Altenberger Straße 69, A-4040 Linz, Austria\\
  \texttt{MalekYacine.Rezoug@oeaw.ac.at}
}

\usepackage{amsmath,amssymb,amsthm,mathtools,mathrsfs,bm}
\usepackage[capitalise,noabbrev]{cleveref}
\usepackage[backend=biber,style=numeric,sorting=nyt]{biblatex}
\usepackage{aliascnt}

\newtheorem{theorem}{Theorem}[section]

\newaliascnt{lemma}{theorem}
\newtheorem{lemma}[lemma]{Lemma}
\aliascntresetthe{lemma}

\newaliascnt{proposition}{theorem}
\newtheorem{proposition}[proposition]{Proposition}
\aliascntresetthe{proposition}

\newaliascnt{corollary}{theorem}

\aliascntresetthe{corollary}

\theoremstyle{definition}

\newaliascnt{definition}{theorem}
\newtheorem{definition}[definition]{Definition}
\aliascntresetthe{definition}

\theoremstyle{remark}

\newaliascnt{remark}{theorem}
\newtheorem{remark}[remark]{Remark}
\aliascntresetthe{remark}

\theoremstyle{definition}

\newaliascnt{assumption}{theorem}
\newtheorem{assumption}[assumption]{Assumption}
\aliascntresetthe{assumption}

\begin{document}

\maketitle

\begin{abstract}
We prove universal approximation for metaplectic neural operators on nonempty
bounded Lipschitz domains. The nonlocal branches are whole-space metaplectic
multipliers with zero extension and restriction, associated with any fixed
free symplectic matrix. Assuming a smooth, globally Lipschitz, nonpolynomial
activation, we obtain compact-uniform approximation of continuous operators
from $C^s$ to $C^r$ and from $H^s$ to $H^r$ in the respective output norms,
for arbitrary nonnegative integers $s,r$ and finite input and output channel
dimensions. In the Sobolev setting, compact input sets must additionally be
uniformly bounded in $L^\infty$. One hidden operator layer with one nonlocal
branch suffices. A single scalar symbol atom and finitely many positional
features are prescribed by the transform and domain independently of the
target operator, compact set and accuracy. The proof combines averaging
neural operator theory with an exact chirp-conjugated averaging identity
and derivative estimates. The result includes Fourier and fixed nondegenerate
fractional Fourier multiplier constructions.
\end{abstract}
\thispagestyle{empty}
 
\section{Introduction}\label{sec:introduction}

Neural operators approximate nonlinear maps between function spaces. A typical
target is the solution operator of a partial differential equation, which maps a
coefficient field, a forcing term or initial data to the corresponding solution.
The dependence on the input need not be local: changing the data in one region
may affect the solution elsewhere. The mathematical analysis of neural operators
must account both for the finite-dimensional nonlinear approximation performed
within the network and for the mechanism by which information is transferred
between different points of the domain. This paper studies the latter mechanism
for neural operators whose nonlocal branches are multipliers with respect to a
prescribed metaplectic transform. We ask whether fixing this transform preserves
universal approximation, including approximation in norms that control
derivatives of the output. Approximation in such norms is relevant when the
predicted field is used to compute gradients or other derivative-dependent
quantities.

A common class of neural operators combines pointwise lifting and projection
maps with nonlinear integral layers~\cite{LiEtAl2020GNO,KovachkiEtAl2023NeuralOperator}.
For a bounded domain $\Omega\subset\mathbb R^d$, the architecture takes the form
$$
\mathcal N
=\mathcal Q\circ\mathcal L_L\circ\cdots\circ\mathcal L_1\circ\mathcal R,
\qquad
\mathcal L_\ell(v)(x)
=\sigma\bigl(W_\ell v(x)+b_\ell(x)+(\mathcal K_\ell v)(x)\bigr),
$$
where $\mathcal R$ and $\mathcal Q$ act pointwise, possibly using spatial
coordinates, $\sigma$ acts componentwise, and $\mathcal K_\ell$ is a linear
nonlocal operator, with dimensions compatible with the channels of the layer.
The lifting maps each input value, together with any positional information,
to the hidden channels, while the projection converts the final hidden state
into the output. The matrix $W_\ell$ and the bias $b_\ell$ act locally; the
nonlocal branch allows the value at $x$ to depend on input values elsewhere in
the domain. Transform-based architectures specify this branch through a chosen
representation. For a multiplier branch, the input is transformed, multiplied
by a symbol, and then mapped back to the spatial variables. The Fourier neural
operator (FNO)~\cite{LiEtAl2021FNO}, for example, multiplies finitely many Fourier
coefficients by trainable matrices, giving a periodic convolution kernel.

Metaplectic analysis provides a common setting for Fourier and fractional
Fourier transforms. Metaplectic operators implement linear symplectic changes
of position and frequency and, up to constant phase factors, include Fourier
transforms, fractional Fourier transforms, dilations and multiplication by
quadratic phases~\cite{Folland1989,deGosson2011,CorderoRodino2020}. Prescribing
such a transform fixes the coordinates in which the multiplier acts. On the
whole space, a metaplectic multiplier need not be translation invariant, even
though a Fourier multiplier is. The free symplectic class considered here
includes linear couplings of position and frequency beyond the scalar rotations
associated with fractional Fourier transforms. This broader class permits a
common approximation question for several choices of transform: which nonlocal
operations are sufficient when the transform itself is fixed?

The classical properties of metaplectic operators describe the linear branch,
but universality also depends on the admissible symbols, positional inputs and
nonlinear maps. We therefore work with prescribed finite families of symbol
atoms and positional features, and with finite neural networks for the local
maps. The objective is to establish universal approximation within this
architecture for every fixed free transform. The construction rests on spatial
averaging, which Lanthaler, Li and Stuart~\cite{LanthalerLiStuart2025} identified
as a sufficient nonlocal operation for universality. Averaging gathers features
of the input over the domain and makes the resulting finite vector available
to the pointwise maps. Here the averaging operation must be realized by the
prescribed metaplectic branch and used in a way compatible with the
componentwise activation.

Let $\Omega\subset\mathbb R^d$ be a nonempty bounded Lipschitz domain, and fix
$$
S=\begin{pmatrix}A&B\\ C&D\end{pmatrix}
\in\operatorname{Sp}(2d,\mathbb R)
$$
that is \emph{free}, meaning that $B$ is invertible. Let $\widehat S$ be the
operator in \eqref{eq:defMetapFree}, which agrees with a metaplectic operator
associated with $S$ up to a constant phase. Using extension by zero
$\mathsf E_\Omega$ and restriction $\mathsf R_\Omega$, the nonlocal branch is
$$
\mathcal K_{\widehat S,P}^{\Omega}
=\mathsf R_\Omega\widehat S^{-1}M_P\widehat S\mathsf E_\Omega,
\qquad (M_Pf)(\eta)=P(\eta)f(\eta),
$$
where $P$ is an essentially bounded matrix-valued symbol and the transforms act
componentwise. This defines a bounded operator on $L^2$, and the constant phase
cancels in the conjugation. Throughout, the whole-space Fourier transform
$\mathscr F$ has the convention in \eqref{eq:Fourier-convention}. The free
metaplectic factorization expresses the branch as a Fourier multiplier
conjugated by multiplication with the quadratic chirp
$$
\gamma_S(x)=e^{\pi i x\cdot Q_Sx},
\qquad Q_S=B^{-1}A.
$$
In particular, symbols in the Schwartz space give convolution kernels multiplied by fixed
quadratic phases in the input and output variables
(\cref{prop:free-transform-conjugation}). The zero-extension and restriction
convention places these branches on the same bounded domain without requiring
periodicity of the input.

The admissible symbols and biases belong to finite-parameter families
$$
P(\eta)=\sum_{j=1}^{M}\Theta_j\rho_j(\eta),
\qquad
b(x)=c+Gp(x),
$$
in which the scalar atoms $\rho_j$ and positional vector $p$ are prescribed,
while the matrix and vector coefficients are trainable. For a fixed free $S$,
our construction uses
$$
p_S(x)=\bigl(x,\operatorname{Re}\gamma_S(x),
                 \operatorname{Im}\gamma_S(x)\bigr).
$$
Each complex channel is represented by two real channels. A metaplectic neural
operator (MpNO) inserts the realified multiplier branches into the layers above
with componentwise real activation. Its lifting and projection are finite
neural networks with positional inputs; the lifting has an activated output,
whereas the projection has an affine output
(\cref{def:finite-direct-metaplectic neural operator}).

Our main result,
\cref{thm:universal-approximation-metaplectic neural operator}, establishes
universality for every fixed free $S$ when the symbol class contains the atom
constructed in \cref{prop:exact-chirped-average} and the positional features
include $p_S$. Assume that $\sigma$ is smooth, globally Lipschitz and
nonpolynomial, as is the case for $\tanh$. For $s,r\in\mathbb N_0$ and
$d_a,d_u\in\mathbb N$, consider either of the settings
$$
\begin{aligned}
X&=C^s(\overline\Omega;\mathbb R^{d_a}),
&Y&=C^r(\overline\Omega;\mathbb R^{d_u}),\\
X&=H^s(\Omega;\mathbb R^{d_a}),
&Y&=H^r(\Omega;\mathbb R^{d_u}).
\end{aligned}
$$
For every continuous $\mathcal G:X\to Y$, every compact $K\subset X$ and every
$\varepsilon>0$, there is a finite MpNO $\mathcal N$ such that
$$
\sup_{a\in K}\|\mathcal G(a)-\mathcal N(a)\|_Y<\varepsilon,
$$
where in the Sobolev setting we additionally assume
$\sup_{a\in K}\|a\|_{L^\infty}<\infty$. For $s>d/2$, Sobolev
embedding~\cite{AdamsFournier2003} makes this bound automatic for compact
subsets of $H^s$; at lower input regularity it remains a hypothesis. The output
norm controls spatial derivatives through order $r$, uniformly in the $C^r$
setting and in $L^2$ in the $H^r$ setting.

In both settings the approximants require only one hidden operator layer with
one nonlocal branch. For each fixed pair $(S,\Omega)$, the transform, a single
scalar symbol atom $\rho_{\widehat S,\Omega}$ and the positional features are
prescribed independently of $\mathcal G$, $K$ and $\varepsilon$. The branch
symbol is a scalar multiple of this atom times the identity matrix. The channel
count, that scalar coefficient, the bias coefficients, and the sizes and weights
of the lifting and projection networks may depend on the approximation data.
Thus the spatial profile of the nonlocal interaction is fixed while the number of features it
communicates may increase. The pointwise networks may have their own depth, and
the fixed atom is a function of the continuous transform variable rather than
a finite selection of Fourier modes.

The proof uses the finite-dimensional reduction of averaging neural operator
theory~\cite{LanthalerLiStuart2025}: the target is approximated by a pointwise
map of finitely many spatial averages and the output position. For every fixed
$(S,\Omega)$, we construct an atom such that, for $q\in\mathbb N$ and
$v\in L^2(\Omega;\mathbb C^q)$,
$$
\mathcal A_Sv
:=\mathcal K^{\Omega}_{\widehat S,\rho_{\widehat S,\Omega}I_q}v
=\overline{\gamma_S}\,\operatorname{Avg}_\Omega(\gamma_Sv),
\qquad
\operatorname{Avg}_\Omega(v)=\frac{1}{|\Omega|}\int_\Omega v(y)\,dy.
$$
The identity holds almost everywhere. It makes the required
averages available through one metaplectic branch. The lifting, bias and
projection are chosen so that this representation can be recovered after the
componentwise activation. Since the spatial dependence after averaging lies in
the fixed smooth chirp, an $L^2$ lifting error yields control of the derivatives
needed for the output norm. Only $L^2$ zero extension is used; no higher
regularity of the zero-extended input is required.

The Fourier and fractional Fourier cases illustrate the result. For
$J=\bigl(\begin{smallmatrix}0&I_d\\-I_d&0\end{smallmatrix}\bigr)$,
one has $\gamma_J=1$ and $\mathcal A_J=\operatorname{Avg}_\Omega$.
For each fixed fractional Fourier angle $\theta$ with $\sin\theta\neq0$, the
associated symplectic rotation is free and has chirp
$e^{\pi i\cot\theta|x|^2}$~\cite{Namias1980,deGosson2011}.
The theorem therefore gives Fourier and fixed-angle fractional Fourier
universality under the same bounded-domain convention and in the stated
derivative norms. These conclusions concern each fixed transform separately;
they do not assert uniform approximation complexity as the transform approaches
a degenerate case.

\subsection{Related work}\label{subsec:related-work}

The approximation of operators by neural networks has a long history. Chen and
Chen~\cite{ChenChen1995} established an early universality theorem, while
DeepONet~\cite{LuEtAl2021DeepONet} and PCA-based
methods~\cite{BhattacharyaEtAl2021PCA} combine finite-dimensional input
representations with finite-dimensional reconstruction of the output. Their
approximation and complexity properties are studied
in~\cite{LanthalerMishraKarniadakis2022,Lanthaler2023PCANet}. Graph-kernel and
multipole graph architectures describe nonlocal interactions directly or through
hierarchical constructions~\cite{LiEtAl2020GNO,LiEtAl2020MGNO}, and a general
framework for neural operators is developed
in~\cite{KovachkiEtAl2023NeuralOperator}.

For FNOs, Kovachki, Lanthaler and
Mishra~\cite{KovachkiLanthalerMishra2021} proved compact-uniform universality
in Sobolev output norms and quantitative error bounds for selected PDE solution
operators. Their nonperiodic formulation uses a bounded Sobolev extension into
a periodic domain followed by restriction. Our construction instead uses
whole-space multipliers with zero extension and restriction, together with a
single prescribed symbol atom. This distinction concerns the realization of the
nonlocal branch; it does not identify the complete nonlinear architectures or
give a quantitative comparison between them. The parametric complexity of
operator learning is studied by Lanthaler and
Stuart~\cite{LanthalerStuart2026Complexity}. Determining whether a particular
metaplectic transform improves approximation rates or reduces network size
requires further information about the target operator class.

The closest approximation result is the ANO theorem of Lanthaler, Li and
Stuart~\cite{LanthalerLiStuart2025}. On bounded Lipschitz domains, it establishes
universality in $C^s$ spaces and in Sobolev
spaces, with a uniform $L^\infty$ input bound in the Sobolev setting. Their
Sobolev result covers the more general $W^{s,p}$ scales. The finite-dimensional
reduction used here is inherited from their proof and is recorded in
\cref{prop:aux-ANO-finite-code-normal-form}. Our contribution is its realization
within the prescribed metaplectic symbol and bias classes, including the
compatibility with activation and the estimates needed to retain the output
norms. In the Fourier case, the construction realizes the ANO average itself as
a whole-space multiplier.

Other transform-based architectures use different representations of the
nonlocal interaction. Multiwavelet operator
learning~\cite{GuptaEtAl2021Multiwavelet} and wavelet neural
operators~\cite{TripuraChakraborty2023WNO} use localized multiscale
representations, while spectral neural
operators~\cite{FanaskovOseledets2022SNO} learn maps between finite vectors of
Fourier or Chebyshev coefficients. Fractional Fourier multiplier constructions
are more directly related to the present work.
CoNO~\cite{TiwariEtAl2025CoNO} uses such branches within a complex-valued
architecture and learns the fractional order in its numerical implementation.
Its stated universality theorem fixes that order and gives compact-uniform
$L^2$ approximation for continuous maps between positive-order Sobolev spaces
on a fractional torus~\cite[Theorem~4.3 in the preprint]{TiwariEtAl2025CoNO}.
Fractional Fourier series on these domains are studied
in~\cite{WangEtAl2024FractionalFourierSeries}. Our fractional Fourier case uses
whole-space transforms on a bounded Lipschitz domain through zero extension and
restriction, and controls $C^r$ or $H^r$ errors under the stated input
hypotheses. The common transform therefore connects the linear branches, while
the domain conventions and nonlinear architectures remain distinct.

Fourier and harmonic-analysis methods provide several approaches to neural
approximation. Barron~\cite{Barron1993} derives approximation bounds for shallow
networks under a weighted integrability condition on the Fourier transform,
while Cand\`es~\cite{Candes1999Harmonic} constructs stable representations by
ridge functions using harmonic-analysis methods. Related approaches use
Fourier symbols~\cite{AbdeljawadDittrich2026Frechet},
time-frequency dictionaries~\cite{AbdeljawadCordero2026TimeFrequency},
and metaplectic dictionaries~\cite{AbdeljawadCarioniCordero2026}.
These works concern the approximation or representation of functions and
linear operators. Here the target is a general continuous nonlinear operator,
approximated using a prescribed metaplectic interaction.

Approximation in derivative norms also connects the present result with
finite-dimensional neural approximation. Simultaneous approximation of functions
and their derivatives is studied
in~\cite{HornikStinchcombeWhite1990,Li1996}. Quantitative Sobolev-norm
estimates are obtained under Fourier-based regularity assumptions and suitable
conditions on the activation in~\cite{SiegelXu2020GeneralActivation},
and for $\tanh$ networks in~\cite{DeRyckLanthalerMishra2021Tanh}.
Further results address Sobolev time-space
approximation~\cite{AbdeljawadGrohs2022Sobolev} and anisotropic
Fourier--Lebesgue and Bochner--Sobolev
settings~\cite{AbdeljawadDittrich2023SpaceTime}. In the present theorem, the
choice of output norm allows the approximation to control derivative-dependent
quantities whenever they define bounded linear maps on the output space.
\subsection{Organization of the paper}

\Cref{sec:preliminaries-direct-metaplectic neural operator} introduces the
function spaces and metaplectic neural operators, establishes the exact
averaging identity and its regularity estimates, and develops the auxiliary
approximation results.
\Cref{sec:universal-approximation-metaplectic neural operator} states and proves
the universal approximation theorem.

\section{Preliminaries and metaplectic neural operators}
	\label{sec:preliminaries-direct-metaplectic neural operator}
	Throughout this paper, $\mathbb N_0=\{0,1,\ldots\}$ and $\mathbb N= \mathbb N_0\setminus \{0\}$. Let $d,d_a,d_u\in\mathbb N$ be fixed,
and $\Omega\subset\mathbb R^d$ is a nonempty bounded Lipschitz domain.
Its Lebesgue measure is denoted by $|\Omega|$.
The vector $\mathbf 1_m\in\mathbb R^m$ has all entries equal to one,
and $M^{\mathsf T}$ denotes the transpose of a matrix $M$.
We write $A\Subset B$ to mean that $\overline{A}$ is compact and $\overline{A}\subset B$,
where the closure is taken in the ambient space.
Standard references for the function spaces used below are
\cite{AdamsFournier2003,RunstSickel1996}; for metaplectic analysis, see
\cite{CorderoNicolaRodino2015,Folland1989,deGosson2011}.

	\subsection{Function spaces and Fourier convention}
	\label{subsec:function-spaces-fourier}
	
	Let $\mathbb F\in\{\mathbb R,\mathbb C\}$ and $q\in\mathbb N$. We write $L^2(\Omega;\mathbb F^q)$ for the set of measurable maps $v:\Omega\to\mathbb F^q$ with
	$$
		\|v\|_{L^2(\Omega;\mathbb F^q)}^2
		:=
		\int_\Omega |v(x)|_{\mathbb F^q}^2\,dx
		<\infty.
	$$
    For a matrix $T$, we denote its Euclidean
	operator norm by
    $\|T\|_{\mathrm{op}} := \sup_{|z|=1}|Tz|$.
    For a multi-index $\alpha=(\alpha_1,\ldots,\alpha_d)\in \mathbb N_0^d$, we
	write $|\alpha|:=\alpha_1+\cdots+\alpha_d$ and
	$\partial^\alpha:=\partial_1^{\alpha_1}\cdots\partial_d^{\alpha_d}$.
	
	Let $m\in \mathbb N_0^d$, the vector-valued Sobolev space is defined by
$$
H^m(\Omega;\mathbb F^q)
:=
\left\{
v\in L^2(\Omega;\mathbb F^q)
:
\partial^\alpha v\in L^2(\Omega;\mathbb F^q)
\text{ for every }\alpha\in\mathbb N_0^d
\text{ with }|\alpha|\le m
\right\},
$$
and is equipped with the norm
$$
\|v\|_{H^m(\Omega;\mathbb F^q)}^2
:=
\sum_{|\alpha|\le m}
\|\partial^\alpha v\|_{L^2(\Omega;\mathbb F^q)}^2.
$$    
    We also use  the vector-valued space $C^m(\overline\Omega;\mathbb F^q)$, consisting of all
$v\in C^m(\Omega;\mathbb F^q)$ such that, for every $\alpha\in\mathbb N_0^d$ with $|\alpha|\le m$,
the derivative $\partial^\alpha v$ admits a continuous extension to $\overline\Omega$.
This space is equipped with the norm
$$
\|v\|_{C^m(\overline\Omega;\mathbb F^q)}
:=
\max_{|\alpha|\le m}
\sup_{x\in\overline\Omega}
|\partial^\alpha v(x)|_{\mathbb F^q}.
$$
For later use, observe that
\begin{equation*}
	\|v\|_{H^m(\Omega;\mathbb F^q)}^2
	\leq
	|\Omega|
	\binom{d+m}{m}
	\|v\|_{C^m(\overline\Omega;\mathbb F^q)}^2,
	\qquad
	v\in C^m(\overline\Omega;\mathbb F^q).
\end{equation*}
Indeed, each term in the definition of the $H^m$ norm is bounded by
$|\Omega|\|v\|_{C^m}^2$, and there are
$\binom{d+m}{m}$ multi-indices $\alpha\in\mathbb N_0^d$ satisfying
$|\alpha|\leq m$. Consequently, $C^m(\overline\Omega;\mathbb F^q)$ is continuously embedded into $H^m(\Omega;\mathbb F^q)$.
The Schwartz space  is defined by 
	$$
		\mathcal S(\mathbb R^d)
		:=
		\left\{
		f\in C^\infty(\mathbb R^d):
		\sup_{x\in\mathbb R^d}
		|x^\alpha\partial^\beta f(x)|<\infty
		\text{ for all multi-indices }\alpha,\beta
		\right\}.
	$$
	It is equipped with the topology generated by the seminorms
$$
p_{\alpha,\beta}(f)
:=
\sup_{x\in\mathbb R^d}
|x^\alpha\partial^\beta f(x)|,
\qquad
\alpha,\beta\in\mathbb N_0^d.
$$
	We use the following Fourier-transform convention
	\begin{equation}
		(\mathscr  F f)(\xi)
		=
		\widehat f(\xi)
		:=
		\int_{\mathbb R^d}
		e^{-2\pi i x\cdot\xi}f(x)\,dx,
		\qquad
		f\in\mathcal S(\mathbb R^d),
		\label{eq:Fourier-convention}
	\end{equation}
	and
	$$
		(\mathscr F^{-1}g)(x)
		:=
		\int_{\mathbb R^d}
		e^{2\pi i x\cdot\xi}g(\xi)\,d\xi,
		\qquad
		g\in\mathcal S(\mathbb R^d).
	$$
	By Plancherel's theorem, $\mathscr F$ extends uniquely to a unitary operator
	on $L^2(\mathbb R^d;\mathbb C)$, and it acts componentwise on
	$L^2(\mathbb R^d;\mathbb C^q)$. For a matrix-valued function
	$\kappa$ and a compatible vector-valued function $f$, convolution means
	$$
		(\kappa*f)(x)
		:=
		\int_{\mathbb R^d}\kappa(x-y)f(y)\,dy
	$$
	whenever the integral is well defined.
	
	For later use, define the bounded linear map
	\begin{equation}
		\operatorname{Avg}_\Omega:
		L^1(\Omega;\mathbb F^q)
		\longrightarrow
		\mathbb F^q,
		\qquad
		\operatorname{Avg}_\Omega(v)
		:=
		\frac{1}{|\Omega|}\int_\Omega v(y)\,dy.
		\label{eq:spatial-average}
	\end{equation}
	The average in \eqref{eq:spatial-average} is a vector in
	$\mathbb F^q$.
	
    \subsection{Symplectic matrices and the metaplectic representation}
    \label{subsec:symplectic-metaplectic}
    We briefly recall the notation and standard facts from symplectic and
    metaplectic analysis that will be used below.  For background, we refer to
    \cite[Chapter~1]{CorderoRodino2020}, \cite[Chapter~4]{Folland1989} and
    \cite{deGosson2011}.
        
    Set
    $$
    	J
    	:=
    	\begin{pmatrix}
    		0&I_d\\
    		-I_d&0
    	\end{pmatrix}.
    $$
    The real symplectic group is
    $$
    	\operatorname{Sp}(2d,\mathbb R)
    	:=
    	\left\{
    	S\in\operatorname{GL}(2d,\mathbb R):
    	S^{\mathsf T}JS=J
    	\right\}.
    $$
    We write
    \begin{equation}
    	S
    	=
    	\begin{pmatrix}
    		A&B\\
    		C&D
    	\end{pmatrix}
    	\in\operatorname{Sp}(2d,\mathbb R),
    	\label{eq:symplectic-blocks}
    \end{equation}
    where $A,B,C,D\in\mathbb R^{d\times d}$.
    Throughout the paper, unless stated otherwise, every
    $S\in\operatorname{Sp}(2d,\mathbb R)$ is written in the block form
    \eqref{eq:symplectic-blocks}, with blocks $A,B,C,D\in\mathbb R^{d\times d}$.
    The identity
    $S^{\mathsf T}JS=J$ implies, among other relations,
\begin{equation*}
A^{\mathsf T}C=C^{\mathsf T}A,
\qquad
B^{\mathsf T}D=D^{\mathsf T}B,
\qquad
A^{\mathsf T}D-C^{\mathsf T}B=I_d,
\qquad
AB^{\mathsf T}=BA^{\mathsf T}.
\end{equation*}


\begin{definition}[Free symplectic matrix]
    		\label{def:free-symplectic}
    		A symplectic matrix $S$ in the block form \eqref{eq:symplectic-blocks} is
    		called \emph{free} if $B$ is invertible.
    	\end{definition}
So for any free symplectic matrix we associate the quadratic function
	\begin{equation*}
			  W_S(x,\xi)
    =
    \frac{1}{2}x^{\mathsf T}DB^{-1}x
    -x^{\mathsf T}B^{-\mathsf T}\xi
    +\frac{1}{2}\xi^{\mathsf T}B^{-1}A\xi.
	\end{equation*} 
To each free symplectic matrix $S$ one can associate Metaplectic operator  given by
	\begin{equation}\label{eq:defMetapFree}
		\widehat{S}f(x):=	\widehat{S}_{W} f(x)
		=
		\frac1{\sqrt{|\det B|}}
		\int_{\mathbb{R}^d}
		e^{2\pi i\, W_S(x,\xi)}
		\,f(\xi)\,d\xi.
	\end{equation}  
The Metaplectic operator is the unitary
operator with an inverse given by
\begin{equation*}
\widehat{S}^{-1} f(x)
=
\frac{1}{\sqrt{|\det B|}}
\int_{\mathbb{R}^{d}}
e^{2\pi i, W_{S^{-1}}(x,\xi)}
f(\xi),d\xi,
\end{equation*} 
such that 
\begin{align*}
W_{S^{-1}}(x,\xi)
&=
-\frac{1}{2} x\cdot A^{\mathsf T}B^{-\mathsf T}x
+ (B^{-1}x)\cdot \xi
-\frac{1}{2}  \xi\cdot B^{-\mathsf T}D^{\mathsf T}\xi
\end{align*}
and
\begin{equation*}
	S^{-1}=
	\begin{pmatrix}
		D^\mathsf T & -B^\mathsf T\\
		- C^\mathsf T & A^\mathsf T
	\end{pmatrix}.
\end{equation*}


    	For a free symplectic matrix, define
    	\begin{equation}
    		Q_S:=B^{-1}A,
    		\qquad
    		T_S:=DB^{-1},
    		\qquad
    		\gamma_S(x):=e^{\pi i x\cdot Q_Sx}.
    		\label{eq:free-matrices-and-chirp}
             \end{equation}
For $f\in\mathcal S(\mathbb R^d)$, we denote
$$
    \begin{aligned}
        (\mathcal C_{T_S}f)(x)
        &:=e^{\pi i\langle T_Sx,x\rangle}f(x),\\
        (\mathcal U_Bf)(x)
        &:=\frac{1}{\sqrt{|\det B|}}\,f(B^{-1}x),\\
        (\mathcal C_{Q_S}f)(x)
        &:=e^{\pi i\langle Q_Sx,x\rangle}f(x),
    \end{aligned}
$$
where $\langle\cdot,\cdot\rangle$ denotes the usual inner product, defined by $\langle u,v\rangle=u^\mathsf Tv$. 

    	Moreover the matrices $Q_S$ and $T_S$ are symmetric. Indeed, the symmetry of $Q_S$
    	follows from $AB^{\mathsf T}=BA^{\mathsf T}$, while the symmetry of $T_S$
    	follows from $B^{\mathsf T}D=D^{\mathsf T}B$.
    	
    \begin{proposition}[Free metaplectic factorization]
    	\label{prop:free-metaplectic-factorization}
    	Let
    	$S	\in\operatorname{Sp}(2d,\mathbb R)
    	$
    	be free. 
        Then
    	\begin{equation*}
    		S
    		=
    		V_{T_S}\Lambda_BJV_{Q_S}.
    	\end{equation*}
where
$$
V_{T_S}:=\begin{pmatrix}I_d&0\\T_S&I_d\end{pmatrix},
\qquad
\Lambda_B:=\begin{pmatrix}B&0\\0&B^{-\mathsf T}\end{pmatrix},
\qquad
J:=\begin{pmatrix}0&I_d\\-I_d&0\end{pmatrix},
$$
and $V_{Q_S}$ is defined analogously, with $T_S$ replaced by $Q_S$.

Moreover, for every $f\in\mathcal S(\mathbb R^d)$,
    \begin{equation}
        \label{eq:free-metaplectic-operator-factorization}
        \begin{aligned}
            \widehat{S}f(x)
            &=
            \bigl(
                \mathcal C_{T_S}
                \mathcal U_B
                \mathscr F
                \mathcal C_{Q_S}f
            \bigr)(x),
        \end{aligned}
    \end{equation}

    \end{proposition}
    
 \begin{proof}
    The symplectic identities
    $B^{\mathsf T}D=D^{\mathsf T}B$ and
    $D^{\mathsf T}A-B^{\mathsf T}C=I$ give
    $C=DB^{-1}A-B^{-\mathsf T}$. Hence
    $$
        V_{T_S}\Lambda_BJV_{Q_S}
        =
        \begin{pmatrix}
            A&B\\
            DB^{-1}A-B^{-\mathsf T}&D
        \end{pmatrix}
        =S.
    $$

    moreover, for $f\in\mathcal S(\mathbb R^d)$,
    $$
        \begin{aligned}
            \bigl(\mathcal C_{T_S}\mathcal U_B
            \mathscr F\mathcal C_{Q_S}f\bigr)(x)&=
            \frac{e^{\pi i x\cdot T_Sx}}{\sqrt{|\det B|}}
            \int_{\mathbb R^d}
            e^{-2\pi i(B^{-1}x)\cdot\xi}
            e^{\pi i\xi\cdot Q_S\xi}f(\xi)\,d\xi\\
            &=
            \frac1{\sqrt{|\det B|}}
            \int_{\mathbb R^d}
            e^{2\pi iW_S(x,\xi)}f(\xi)\,d\xi
            =\widehat{S}f(x).
        \end{aligned}
    $$
\end{proof}
	
	\subsection{Bounded-domain metaplectic multipliers}
	\label{subsec:bounded-domain-metaplectic-multipliers}
	
	For $q\in\mathbb N$, define extension by zero and restriction by
	\begin{align*}
		\mathsf E_{\Omega,q}:
		L^2(\Omega;\mathbb C^q)
		&\longrightarrow
		L^2(\mathbb R^d;\mathbb C^q),
		&
		(\mathsf E_{\Omega,q}v)(x)
		&:=
		\begin{cases}
			v(x),&x\in\Omega,\\
			0,&x\notin\Omega,
		\end{cases}
		\\
		\mathsf R_{\Omega,q}:
		L^2(\mathbb R^d;\mathbb C^q)
		&\longrightarrow
		L^2(\Omega;\mathbb C^q),
		&
		\mathsf R_{\Omega,q}f
		&:=f|_\Omega.
	\end{align*}
These operators are bounded and satisfy the following estimates:
\begin{equation}
    \|\mathsf E_{\Omega,q}v\|_{L^2(\mathbb R^d;\mathbb C^q)}
    =
    \|v\|_{L^2(\Omega;\mathbb C^q)},
    \qquad
    \|\mathsf R_{\Omega,q}f\|_{L^2(\Omega;\mathbb C^q)}
    \leq
    \|f\|_{L^2(\mathbb R^d;\mathbb C^q)}.
    \label{eq:extension-restriction-bounds}
\end{equation}

    For $q\in\mathbb N$, the componentwise metaplectic operator is
    $
    \widehat{S}_q: L^2(\mathbb R^d;\mathbb C^q)
    \longrightarrow
    L^2(\mathbb R^d;\mathbb C^q)$
    such that
    $$
   \widehat{S}_q(f):=\widehat{S}_q(f_1,\ldots,f_q)
    :=
    \bigl(\widehat{S}f_1,\ldots,\widehat{S}f_q\bigr),
    \qquad f_j\in L^2(\mathbb R^d).
    $$
	Note that $ \widehat{S}_q$ is unitary. Let $q_-,q_+\in\mathbb N$ and let
	\begin{equation}
		P:\mathbb R^d\longrightarrow
		\mathbb C^{q_+\times q_-}
		\label{eq:matrix-symbol}
	\end{equation}
	be a measurable and essentially bounded matrix-valued function. We define the associated multiplication operator
	\begin{equation}
		M_P:
		L^2(\mathbb R^d;\mathbb C^{q_-})
		\longrightarrow
		L^2(\mathbb R^d;\mathbb C^{q_+}),
		\qquad
		(M_Pg)(\xi):=P(\xi)g(\xi).
		\label{eq:matrix-multiplier}
	\end{equation}
    The essential boundedness of $P$ guarantees that $M_P$ is a well-defined bounded linear operator. For a
    matrix-valued symbol $P$ we set
    \begin{equation*}
	\|P\|_{L^\infty_{\mathrm{op}}}
	:=
	\operatorname*{ess\,sup}_{\xi\in\mathbb R^d}
	\|P(\xi)\|_{\mathrm{op}}.
\end{equation*}

    \begin{lemma}[Norm of a matrix-valued multiplication operator]
	\label{lem:matrix-multiplier-norm}
	The operator in \eqref{eq:matrix-multiplier} is bounded and
	\begin{equation}
		\|M_P\|_{L^2\to L^2}
		=
		\|P\|_{L^\infty_{\mathrm{op}}}.
		\label{eq:matrix-multiplier-exact-norm}
	\end{equation}
    \end{lemma}

\begin{proof}
The proof is similar to \cite[Theorem 1.1]{Hudzik06MatrixMultiplicationOperators}. Therefore, the details are left to the reader.
\end{proof}

	\begin{definition}
		\label{def:bounded-domain-direct-metaplectic-multiplier}
		For $S$ free symplectic matrix,  and let $P$ be a matrix-valued symbol as in \cref{eq:matrix-symbol}.
The associated  \emph{kernel branch}  $$
\mathcal K_{\widehat S,P}^{\Omega}:
L^2(\Omega;\mathbb C^{q_-})
\longrightarrow
L^2(\Omega;\mathbb C^{q_+})
$$ is defined  as following 
\begin{equation*}
\label{eq:branch-kernel}
\mathcal K_{\widehat S,P}^{\Omega}
:=
\mathsf R_{\Omega,q_+}
\widehat S_{q_+}^{-1}
M_P
\widehat S_{q_-}
\mathsf E_{\Omega,q_-}.
\end{equation*}
\end{definition}
\begin{remark}
\label{remark:bounded-domain-metaplectic-bound}
From \eqref{eq:extension-restriction-bounds}, and
	\eqref{eq:matrix-multiplier-exact-norm} , we  deduce that  is the bounded linear operator  and satisfies the following inequality 
		$$
			\|\mathcal K_{\widehat S,P}^{\Omega}\|_{L^2\to L^2}
			\leq
			\|P\|_{L^\infty_{\mathrm{op}}}.
		$$
\end{remark}
	
The following proposition shows that, for a free symplectic
matrix, the branch kernel can be represented as a
chirp-conjugated Fourier multiplier. 
\begin{proposition}[Free-transform conjugation formula]
	\label{prop:free-transform-conjugation}
	Let $S$ be free, 
    and let $P$ be as
	in \eqref{eq:matrix-symbol}. 
    
    Then 
	\begin{equation}
		\widehat{S}_{q_+}^{-1}M_P\widehat{S}_{q_-}
		=
		\mathcal C_{-Q_S}
		\mathscr F^{-1}
		M_{P(B\,\cdot)}
		\mathscr F
		\mathcal C_{Q_S},
		\label{eq:free-conjugated-multiplier}
	\end{equation}
	moreover, 
	\begin{equation}
		\mathcal K_{\widehat S,P}^{\Omega}
		=
		\mathsf R_{\Omega,q_+}
		\mathcal C_{-Q_S}
		\mathscr F^{-1}
		M_{P(B\,\cdot)}
		\mathscr F
		\mathcal C_{Q_S}
		\mathsf E_{\Omega,q_-}.
		\label{eq:bounded-free-conjugated-multiplier}
	\end{equation}
\end{proposition}

\begin{proof}
	Combining \cref{eq:free-metaplectic-operator-factorization} and the identities 
    $\mathcal C_{-T_S}M_P\mathcal C_{T_S} = M_P$
    and
    $\mathcal U_B^{-1}M_P\mathcal U_B
	=
	M_{P(B\,\cdot)}$,  we get \cref{eq:free-conjugated-multiplier}. \cref{def:bounded-domain-direct-metaplectic-multiplier} and \cref{eq:free-conjugated-multiplier}
    give \eqref{eq:bounded-free-conjugated-multiplier}.
\end{proof}
This formula will be used below (see \cref{prop:exact-chirped-average}) with a scalar-valued symbol for which the
Fourier multiplier becomes convolution with a kernel that is constant on $\Delta_{\Omega}$ from \eqref{eq:delta-omega}.

        \subsection{Realification and finite-parameter classes}
\label{subsec:realification-finite-parameters}

The metaplectic transform is complex-valued even when its input is real. We
therefore encode each complex channel by two real channels.

\begin{definition}[Realification]
	  \label{def:realification}
    For $q\in\mathbb N$, the map
    $$
        \mathfrak r_q:\mathbb C^q\longrightarrow\mathbb R^{2q},
        \qquad
        \mathfrak r_q(z)
        :=
        \begin{pmatrix}
            \operatorname{Re}z\\
            \operatorname{Im}z
        \end{pmatrix}
    $$
    is called the realification map.

   \end{definition}
   \begin{remark}
        For $\Theta=U+iV\in\mathbb C^{q_+\times q_-}$, define
    $$
        [\Theta]_{\mathbb R}
        :=
        \begin{pmatrix}
            U&-V\\
            V&U
        \end{pmatrix}
        \in\mathbb R^{2q_+\times2q_-}.
    $$
    We also call $\Theta\mapsto[\Theta]_{\mathbb R}$  \emph{realification}.
    
    For every $z\in\mathbb C^{q_-}$,
    $$
        \mathfrak r_{q_+}(\Theta z)
        =
        [\Theta]_{\mathbb R}\mathfrak r_{q_-}(z).
    $$
   \end{remark}
   
\begin{remark}
    Realification function commutes with differentiation and preserves the   standard $C^r$ and $H^r$ norms. For any bounded complex-linear  operator $T:L^2(\Omega;\mathbb C^{q_-})\to
    L^2(\Omega;\mathbb C^{q_+})$, its realification
    $$
        T_{\mathbb R}
        :=\mathfrak r_{q_+}T\mathfrak r_{q_-}^{-1}
    $$
    is real-linear and satisfies  $\|T_{\mathbb R}\|_{\mathrm{op}}=\|T\|_{\mathrm{op}}$.
\end{remark}

We can now define the realified bounded-domain metaplectic multiplier by \begin{equation}
	\mathbf K_{\widehat S,P}^{\Omega}
	:=
	\mathfrak r_{q_+}
	\mathcal K_{\widehat S,P}^{\Omega}
	\mathfrak r_{q_-}^{-1}.
	\label{eq:realified-bounded-domain-metaplectic-multiplier}
\end{equation}
\begin{definition}[Finite-parameter symbol class]
	\label{def:finite-parameter-symbol-class}
	Fix $q_-,q_+,M\in\mathbb N$ and fixed  $\rho_1,\ldots,\rho_M	\in L^\infty(\mathbb R^d;\mathbb C).	$
	For trainable matrices $\Theta_1, \dots,\Theta_M\in\mathbb C^{q_+\times q_-}$, define
    $$
		P_\theta(\xi)
		:=
		\sum_{r=1}^M\Theta_r\rho_r(\xi),
		\qquad
		\theta=(\Theta_1,\ldots,\Theta_M).
	$$
\end{definition}

The admissible trainable biases are restricted to the following
finite-dimensional function space.

\begin{definition}[Finite-parameter bias class]
	\label{def:finite-parameter-spatial-bias}	Fix $m\in\mathbb N_0$ and a fixed positional vector
	$$
		p=(p_1,\ldots,p_m):
		\overline\Omega\longrightarrow\mathbb R^m,
		\qquad
		p_j\in C^\infty(\overline\Omega).
	$$
	For $n\in\mathbb N$, a trainable bias function has the form
	$$
		b(x)
		:=
		c+Gp(x),
		\qquad
		c\in\mathbb R^n,
		\quad
		G\in\mathbb R^{n\times m}.
	$$
	When $m=0$, or when $G=0$, this reduces to a constant bias.
\end{definition}

For a fixed free transform, a useful positional vector is
\begin{equation}
	p_S(x)
	:=
	\bigl(
	x_1,\ldots,x_d,
	\cos(\pi x\cdot Q_Sx),
	\sin(\pi x\cdot Q_Sx)
	\bigr).
	\label{eq:canonical-metaplectic-position}
\end{equation}
This allows real-valued networks to use the spatial coordinates and the
real and imaginary parts of the fixed chirp $\gamma_S$ as inputs.
The architecture may use any fixed finite family of positional features
containing those required in the proof; these features are prescribed
and are not trainable. 
\begin{remark}
    More general formulations are possible in which the symbol and the 
    bias function are allowed to vary in infinite-dimensional function classes.
    We do not pursue such generalizations here, since the finite-dimensional
    classes introduced above are sufficient for all constructions in this paper.
\end{remark}

\subsection{Activation and local neural maps}
\label{subsec:activation-local-maps}

\begin{assumption}[Activation]
	\label{ass:activation}
	The activation  function $\sigma:\mathbb R\to\mathbb R$ is nonpolynomial, globally
	Lipschitz, and  smooth. Its Lipschitz constant is denoted by $L_\sigma$. 
\end{assumption}
For $n\in\mathbb N$, denote the component wise map by
$$
	\Sigma_n(z_1,\ldots,z_n)
	:=
	\bigl(\sigma(z_1),\ldots,\sigma(z_n)\bigr).
$$

\begin{lemma}[A locally invertible activation interval]
	\label{lem:local-inverse-activation}
	There exist $t_*\in\mathbb R$ and open intervals
	$I_\sigma,J_\sigma\subset\mathbb R$ such that $t_*\in I_\sigma$, $\sigma(t_*)\in J_\sigma$, and
	$$
		\sigma|_{I_\sigma}:I_\sigma\longrightarrow J_\sigma
	$$
	is a $C^\infty$ diffeomorphism.
\end{lemma}

\begin{proof}
  The proof follows from applying the inverse function theorem.
 \end{proof}
We next fix a finite-dimensional neural-network convention.  Let $N\in\mathbb N$ and
$n_0,\ldots,n_{N}\in\mathbb N$.  An
\emph{activated-output} network is defined by
$$
	z_0=z,
	\qquad
	z_j
	=
	\Sigma_{n_j}(A_jz_{j-1}+a_j),
	\quad
	j=1,\ldots,N,
	\qquad
	\Phi(z):=z_{N}.
$$
An \emph{affine-output} network is defined by the same recursion for
$j=1,\ldots,N-1$ and by
$$
	\Phi(z)
	:=
	A_{N}z_{N-1}
	+a_{N}.
$$

Since the scalar activation $\sigma$ is globally Lipschitz with constant
$L_\sigma$, an activated-output network is globally Lipschitz with
constant at most
$L_\sigma^{N}
	\prod_{j=1}^{N}
	\|A_j\|_{\mathrm{op}}.$
For an affine-output network, the corresponding bound is
$	\|A_{N}\|_{\mathrm{op}}
	L_\sigma^{N-1}
	\prod_{j=1}^{N-1}
	\|A_j\|_{\mathrm{op}}.
$

Fix
$q_{\mathrm{in}},q_{\mathrm{out}}\in\mathbb N$ and
$m_{\mathrm{in}},m_{\mathrm{out}}\in\mathbb N_0$, and let
$
	p_{\mathrm{in}}
	\in
	C^\infty(\overline\Omega;\mathbb R^{m_{\mathrm{in}}}),
	$
	$
    p_{\mathrm{out}}
	\in
	C^\infty(\overline\Omega;\mathbb R^{m_{\mathrm{out}}})
$
be fixed positional vectors.

Let
$\Phi^{\mathrm{lift}}$ be an
activated-output network from
$\mathbb R^{d_a+m_{\mathrm{in}}}$ to
$\mathbb R^{2q_{\mathrm{in}}}$, and let
$\Phi^{\mathrm{out}}$ be an affine-output
network from
$\mathbb R^{2q_{\mathrm{out}}+m_{\mathrm{out}}}$ to
$\mathbb R^{d_u}$.  Their pointwise realizations are
\begin{align}
	\mathcal R:
	L^2(\Omega;\mathbb R^{d_a})
	&\longrightarrow
	L^2(\Omega;\mathbb R^{2q_{\mathrm{in}}}),
	&
	\mathcal R(a)(x)
	&:=
	\Phi^{\mathrm{lift}}
	\bigl(a(x),p_{\mathrm{in}}(x)\bigr),
	\label{eq:local-lifting-map}
	\\
	\mathcal Q:
	L^2(\Omega;\mathbb R^{2q_{\mathrm{out}}})
	&\longrightarrow
	L^2(\Omega;\mathbb R^{d_u}),
	&
	\mathcal Q(v)(x)
	&:=
	\Phi^{\mathrm{out}}
	\bigl(v(x),p_{\mathrm{out}}(x)\bigr).
	\label{eq:local-readout-map}
\end{align}

The values $m_{\mathrm{in}}=0$ or $m_{\mathrm{out}}=0$ are allowed.  For
example, when $m_{\mathrm{in}}=0$, the lifting network depends only on
$a(x)$ and receives no additional positional information.

Since $p_{\mathrm{in}}$ and $p_{\mathrm{out}}$ are bounded on
$\overline\Omega$ and $|\Omega|<\infty$, the global Lipschitz continuity
of $\Phi^{\mathrm{lift}}$ and $\Phi^{\mathrm{out}}$ implies that the
pointwise maps $\mathcal R$ and $\mathcal Q$ are well defined between
the corresponding $L^2$ spaces.  With the positional inputs fixed, the same
Lipschitz estimates show that $\mathcal R$ and $\mathcal Q$ are globally
Lipschitz with respect to their function-valued inputs.

The activated-output convention in \eqref{eq:local-lifting-map} will be convenient when composing
the maps with the metaplectic layers. The final output map is affine-output,
so no activation is applied at the last layer, and its values are not
constrained to lie in the range of $\sigma$. Moreover, an internal hidden
layer may always be enlarged by adding unused units and setting all
outgoing weights from those units equal to zero. Thus, when needed for
rectification, its width may be assumed to be even.

\begin{definition}[Finite 
metaplectic neural operator]
	\label{def:finite-direct-metaplectic neural operator}
	Let $L\in\mathbb N$ and let
	$q_0,\ldots,q_L\in\mathbb N$.  Let
	$$
		\mathcal R:
		L^2(\Omega;\mathbb R^{d_a})
		\longrightarrow
		L^2(\Omega;\mathbb R^{2q_0})
	$$
	be a lifting  operator of the form
	\eqref{eq:local-lifting-map}, and let
	$$
		\mathcal Q:
		L^2(\Omega;\mathbb R^{2q_L})
		\longrightarrow
		L^2(\Omega;\mathbb R^{d_u})
	$$
	be a projection operator of the form
	\eqref{eq:local-readout-map}.

	For each $\ell\in\{1,\ldots,L\}$, choose
	$J_\ell\in\mathbb N_0$, a matrix
		$W_\ell
		\in
		\mathbb R^{2q_\ell\times2q_{\ell-1}}$,
	and a finite-parameter bias function
	$b_\ell:
		\overline\Omega
		\longrightarrow
		\mathbb R^{2q_\ell}$
	as in Definition~\ref{def:finite-parameter-spatial-bias}.
	For every $j\in\{1,\ldots,J_\ell\}$, choose a Symplectic
	$S_{\ell,j}$ and a symbol
	$$
		P_{\ell,j}:
		\mathbb R^d
		\longrightarrow
		\mathbb C^{q_\ell\times q_{\ell-1}}
	$$
	belonging to a finite-parameter symbol class as in
	Definition~\ref{def:finite-parameter-symbol-class}.
Let
	$\mathcal K_{\widehat S_{\ell,j},P_{\ell,j}}^\Omega$
	be the associated bounded-domain metaplectic multiplier from
	Definition~\ref{def:bounded-domain-direct-metaplectic-multiplier}.
    
	Its realification
	$\mathbf K_{\widehat S_{\ell,j},P_{\ell,j}}^\Omega$
	is defined by
	\eqref{eq:realified-bounded-domain-metaplectic-multiplier}.
	The corresponding hidden metaplectic layer is
	$$
		\mathcal L_\ell:
		L^2(\Omega;\mathbb R^{2q_{\ell-1}})
		\longrightarrow
		L^2(\Omega;\mathbb R^{2q_\ell}),
	$$
	defined by
	\begin{equation}
		\mathcal L_\ell(v)(x)
		:=
		\Sigma_{2q_\ell}
		\left(
			W_\ell v(x)
			+
			b_\ell(x)
			+
			\sum_{j=1}^{J_\ell}
			\left(
				\mathbf K_{\widehat S_{\ell,j},P_{\ell,j}}^\Omega
				v
			\right)(x)
		\right).
		\label{eq:finite-direct-metaplectic neural operator-layer}
	\end{equation}
	When $J_\ell=0$, the sum in
	\eqref{eq:finite-direct-metaplectic neural operator-layer} is understood to be zero.

	The operator
	\begin{equation}
		\mathcal N
		:=
		\mathcal Q
		\circ
		\mathcal L_L
		\circ\cdots\circ
		\mathcal L_1
		\circ
		\mathcal R
		\label{eq:finite-direct-metaplectic neural operator}
	\end{equation}
	is called a \emph{finite metaplectic neural operator}.
	If $L=1$, then $\mathcal N$ is said to have one hidden metaplectic layer.
\end{definition}

\begin{lemma}[Compact range encoding with an activated output]
	\label{lem:activated-output-range-encoding}
	Let $K\subset\mathbb R^n$ be compact and let
	$f:K\to\mathbb R^m$ be continuous.  There exist
	$a_f\in\mathbb R^m$ and $\tau_f>0$ such that, for every
	$\varepsilon>0$, an activated-output network $\Phi$ can be chosen
	with
	\begin{equation}
		\sup_{z\in K}
		\bigl|
		\Phi(z)-a_f-\tau_f f(z)
		\bigr|
		<
		\varepsilon.
		\label{eq:activated-output-range-encoding}
	\end{equation}
    
	If, in addition, $r\in\mathbb N_0$, the compact set $K$ is contained
	in an open set $U\subset\mathbb R^n$, and
	$f\in C^r(U;\mathbb R^m)$, then $a_f$ and $\tau_f$ can be chosen so
	that, for every $\varepsilon>0$, the network $\Phi$ satisfies
	\begin{equation}
		\max_{|\alpha|\leq r}
		\sup_{z\in K}
		\left|
		\partial^\alpha\Phi(z)
		-
		\partial^\alpha
		\bigl(a_f+\tau_f f(z)\bigr)
		\right|
		<
		\varepsilon.
		\label{eq:activated-output-Cr-range-encoding}
	\end{equation}
\end{lemma}

\begin{proof}
	Let $t_*$ and the intervals $I_\sigma,J_\sigma$ be as in
	\cref{lem:local-inverse-activation}. 
    Choose $\rho>0$ such that
	$
		[\sigma(t_*)-\rho,\sigma(t_*)+\rho]
		\subset
		J_\sigma.
	$
	Note that $f(K)$ is a bounded set .  Hence there exists
	$M_f\geq0$ such that
	$$
		|f_j(z)|
		\leq
		M_f,
		\qquad
		z\in K,
		\quad
		j=1,\ldots,m.
	$$
	Set	$
		a_f
		:=
		\sigma(t_*)\mathbf 1_m
	$
	and choose
	$
		0<\tau_f
		<
		\frac{\rho}{1+M_f}.
	$
	Then
	$
		|\tau_f f_j(z)|
		<
		\rho
	$
	for every $z\in K$ and $j=1,\ldots,m$.  Consequently, every
	component of $a_f+\tau_f f(z)$ belongs to $J_\sigma$.

	The inverse of $\sigma|_{I_\sigma}$ can therefore be applied
	componentwise, and the map
	$$
		g(z)
		:=
		(\sigma|_{I_\sigma})^{-1}
		\bigl(a_f+\tau_f f(z)\bigr),
		\qquad
		z\in K,
	$$
	is well defined and continuous.  By the finite-dimensional universal
	approximation theorem
	\cite{LeshnoLinPinkusSchocken1993}, there exists an affine-output
	network $\Psi$ that approximates $g$ uniformly on $K$.  The
	vector-valued case is obtained by applying the scalar approximation
	result to each component of $g$.

	Append one componentwise activation and define
	$$
		\Phi
		:=
		\Sigma_m\circ\Psi.
	$$
	This is an activated-output network.  By the definition of $g$,
	$
		\Sigma_m(g(z))
		=
		a_f+\tau_f f(z).
	$
	Then 
    $$
		|\Phi(z)-a_f-\tau_f f(z)|
		=
		|\Sigma_m(\Psi(z))-\Sigma_m(g(z))|
		\leq
		L_\sigma|\Psi(z)-g(z)|.
	$$
	Thus, by approximating $g$ sufficiently accurately with $\Psi$, we
	obtain \eqref{eq:activated-output-range-encoding}.

	For the $C^r$ statement, choose a compact neighborhood
	$K_1\subset U$ such that
	$
		K
		\subset
		\operatorname{int}(K_1).
	$
	Since $f$ is continuous, it is bounded on $K_1$.  We choose
	$a_f=\sigma(t_*)\mathbf 1_m$ as above and choose $\tau_f>0$ using
	a bound for $f$ on $K_1$.  Then every component of
	$a_f+\tau_f f(z)$ lies in $J_\sigma$ for every $z$ in a
	neighborhood of $K$.  Hence
	$$
		g(z)
		=
		(\sigma|_{I_\sigma})^{-1}
		\bigl(a_f+\tau_f f(z)\bigr)
	$$
	is of class $C^r$ on a neighborhood of $K$.

	By the simultaneous approximation theorem for functions and their
	derivatives \cite[Theorem~2.1]{Li1996}, the affine-output network
	$\Psi$ can be chosen so that $g$ and all its derivatives up to order
	$r$ are approximated uniformly on $K$.  Since $\sigma$ is smooth,
	the chain rule implies that
	$$
		\Sigma_m\circ\Psi
		\longrightarrow
		\Sigma_m\circ g
		=
		a_f+\tau_f f
	$$
	in the $C^r$ norm on $K$.  Choosing the approximation sufficiently
	accurate gives
	\eqref{eq:activated-output-Cr-range-encoding}.
\end{proof}

\subsection{Exact metaplectic averaging branch}
\label{subsec:exact-metaplectic-averaging}

We now construct the single fixed symbol atom used in the later
universality proof. Fix a free symplectic matrix $S$. Define the compact symmetric difference set
\begin{equation}\label{eq:delta-omega}
    \Delta_\Omega
    :=
    \overline{\Omega} - \overline{\Omega}
    =
    \{x-y : x,y \in \overline{\Omega}\}.
\end{equation}
Choose a real-valued even function
\begin{equation}
	\chi_\Omega\in C_c^\infty(\mathbb R^d),
	\qquad
	\chi_\Omega(z)=1
	\quad
	\text{on an open neighborhood of }\Delta_\Omega.
	\label{eq:difference-set-cutoff}
\end{equation}
Set
\begin{equation}
	\kappa_\Omega(z)
	:=
	\frac{1}{|\Omega|}\chi_\Omega(z),
	\qquad
	\rho_{\widehat S,\Omega}(\eta)
	:=
	\widehat{\kappa_\Omega}(B^{-1}\eta).
	\label{eq:metaplectic-averaging-symbol}
\end{equation}

\begin{proposition}[Exact chirp-conjugated averaging]
	\label{prop:exact-chirped-average}
	For every $q\in\mathbb N$ and every
$v\in L^2(\Omega;\mathbb C^q)$,
\begin{equation}
    \mathcal K_{\widehat S,
        \rho_{\widehat S,\Omega}I_q}^{\Omega}v(x)
    =
    \overline{\gamma_S(x)}
    \operatorname{Avg}_\Omega(\gamma_Sv)
    \label{eq:exact-chirped-average}
\end{equation}
for almost every $x\in\Omega$, where $I_q$ denotes the
$q\times q$ identity matrix.

More generally, if $q_-,q_+\in\mathbb N$,
	$\Theta\in\mathbb C^{q_+\times q_-}$, and
	\begin{equation}
		P_{\Theta}(\eta)
		:=
		\Theta\rho_{\widehat S,\Omega}(\eta),
		\label{eq:matrix-averaging-symbol}
	\end{equation}
	then
	\begin{equation}
		\mathcal K_{\widehat S,P_\Theta}^{\Omega}v(x)
		=
		\overline{\gamma_S(x)}
		\Theta
		\operatorname{Avg}_\Omega(\gamma_Sv)
		\label{eq:exact-matrix-chirped-average}
	\end{equation}
	for every $v\in L^2(\Omega;\mathbb C^{q_-})$.
\end{proposition}

\begin{proof}
	By the definition of $\rho_{\widehat S,\Omega}$ in
	\eqref{eq:metaplectic-averaging-symbol}, one has
	\begin{equation}
		\rho_{\widehat S,\Omega}(B\xi)
		=
		\widehat{\kappa_\Omega}(B^{-1}B\xi)
		=
		\widehat{\kappa_\Omega}(\xi).
		\label{eq:averaging-symbol-rescaling}
	\end{equation}
	
	We first consider the symbol
	$\rho_{\widehat S,\Omega}I_q$. By
	Proposition~\ref{prop:free-transform-conjugation},
	\begin{align*}
		\mathcal K_{\widehat S,
			\rho_{\widehat S,\Omega}I_q}^{\Omega}v
		=
		\mathsf R_{\Omega,q}
		\left(
		\overline{\gamma_S}\,
		\mathscr F^{-1}
		\left[
		\rho_{\widehat S,\Omega}(B\cdot)I_q\,
		\mathscr F
		\bigl(
		\mathsf E_{\Omega,q}(\gamma_Sv)
		\bigr)
		\right]
		\right).
	\end{align*}
	Using \eqref{eq:averaging-symbol-rescaling}, this becomes
	\begin{align*}
		\mathcal K_{\widehat S,
			\rho_{\widehat S,\Omega}I_q}^{\Omega}v
		=
		\mathsf R_{\Omega,q}
		\left(
		\overline{\gamma_S}\,
		\mathscr F^{-1}
		\left[
		\widehat{\kappa_\Omega}I_q\,
		\mathscr F
		\bigl(
		\mathsf E_{\Omega,q}(\gamma_Sv)
		\bigr)
		\right]
		\right).
	\end{align*}
	
	The Fourier transform is applied componentwise to
	$\mathbb C^q$-valued functions.
	Therefore, for almost every $x\in\Omega$,
	\begin{align}
		\mathcal K_{\widehat S,
			\rho_{\widehat S,\Omega}I_q}^{\Omega}v(x)
		&=
		\overline{\gamma_S(x)}
		\bigl[
		\kappa_\Omega*
		\mathsf E_{\Omega,q}(\gamma_Sv)
		\bigr](x)
		\notag\\
		&=
		\overline{\gamma_S(x)}
		\int_{\mathbb R^d}
		\kappa_\Omega(x-y)
		\mathsf E_{\Omega,q}(\gamma_Sv)(y)\,dy
		\notag\\
		&=
		\overline{\gamma_S(x)}
		\int_\Omega
		\kappa_\Omega(x-y)\gamma_S(y)v(y)\,dy.
		\label{eq:chirped-convolution-calculation}
	\end{align}
	The last equality follows from the definition of the zero extension
	$\mathsf E_{\Omega,q}$.
	
	For $x,y\in\Omega$, one has
	$
	x-y\in\Omega-\Omega\subseteq\Delta_\Omega.
	$
	By \eqref{eq:difference-set-cutoff},
	$\chi_\Omega(x-y)=1$, and hence
	$$
	\kappa_\Omega(x-y)
	=
	\frac{1}{|\Omega|}.
	$$
	Substituting this identity into
	\eqref{eq:chirped-convolution-calculation}, we obtain
	$$
\mathcal K_{\widehat S,\rho_{\widehat S,\Omega}I_q}^{\Omega}v(x) = \overline{\gamma_S(x)}\frac{1}{|\Omega|}\int_\Omega \gamma_S(y)v(y)\,dy = \overline{\gamma_S(x)}\operatorname{Avg}_\Omega(\gamma_Sv).
$$ 

	The integral and the average are understood componentwise and therefore
	belong to $\mathbb C^q$. This proves
	\eqref{eq:exact-chirped-average}.
	
	For the second assertion, since
	$
	P_\Theta
	=
	\Theta\bigl(\rho_{\widehat S,\Omega}I_{q_-}\bigr)
	$
	and $\Theta$ is constant and  $I_{q_-}$ is  the $q_-\times q_-$  identity matrix, yields 
	$$
	\mathcal K_{\widehat S,P_\Theta}^{\Omega}v
	=
	\Theta
	\mathcal K_{\widehat S,
		\rho_{\widehat S,\Omega}I_{q_-}}^{\Omega}v.
	$$
	Hence \eqref{eq:exact-chirped-average}, applied with $q=q_-$, yields
	$$
	\mathcal K_{\widehat S,P_\Theta}^{\Omega}v(x)
	=
	\overline{\gamma_S(x)}
	\Theta\operatorname{Avg}_\Omega(\gamma_Sv)
	$$
	for almost every $x\in\Omega$, proving
	\eqref{eq:exact-matrix-chirped-average}.
\end{proof}

\begin{proposition}[Finite rank and positive regularity of the averaging branch]
		\label{prop:averaging-branch-regularity}
	Let $r\in\mathbb N_0$, let $\Theta\in\mathbb C^{q_+\times q_-}$,
and let $P_\Theta$ be defined by
\eqref{eq:matrix-averaging-symbol}. The range of
$\mathcal K_{\widehat S,P_\Theta}^{\Omega}$ acting on
$L^2(\Omega;\mathbb C^{q_-})$ is contained in the
finite-dimensional space
$$
\mathcal V_{S,q_+}
:=
\left\{
x\longmapsto\overline{\gamma_S(x)}c:
c\in\mathbb C^{q_+}
\right\}.
$$
Moreover, for every $v\in L^2(\Omega;\mathbb C^{q_-})$,
\begin{align}
    \|\mathcal K_{\widehat S,P_\Theta}^{\Omega}v\|_{C^r(\Omega;\mathbb C^{q_+})}
    &\leq
    C^{C}_{r,S,\Omega}
    \|\Theta\|_{\mathrm{op}}
    |\Omega|^{-1/2}
    \|v\|_{L^2(\Omega;\mathbb C^{q_-})},
    \label{eq:averaging-branch-Cr-bound}
    \\
    \|\mathcal K_{\widehat S,P_\Theta}^{\Omega}v\|_{H^r(\Omega;\mathbb C^{q_+})}
    &\leq
    C^{H}_{r,S,\Omega}
    \|\Theta\|_{\mathrm{op}}
    |\Omega|^{-1/2}
    \|v\|_{L^2(\Omega;\mathbb C^{q_-})},
    \label{eq:averaging-branch-Hr-bound}
\end{align}
where
$$
C^{C}_{r,S,\Omega} := \max_{|\alpha|\leq r}\|\partial^\alpha\gamma_S\|_{L^\infty(\Omega;\mathbb C)}, \qquad C^{H}_{r,S,\Omega} := \left(\sum_{|\alpha|\leq r}\|\partial^\alpha\gamma_S\|_{L^2(\Omega;\mathbb C)}^2\right)^{1/2}.
$$ 
	\end{proposition}
	
	\begin{proof}
		The range statement follows directly from
		\eqref{eq:exact-matrix-chirped-average}. Since $|\gamma_S|=1$, the fact that by Cauchy--Schwarz,
    	\begin{equation}
    		|\operatorname{Avg}_\Omega(v)|
    		\leq
    		|\Omega|^{-1/2}\|v\|_{L^2(\Omega)}, 
    	\end{equation}
		we get
		\begin{equation}
			\left|
			\Theta\operatorname{Avg}_\Omega(\gamma_Sv)
			\right|
			\leq
			\|\Theta\|_{\mathrm{op}}
			|\Omega|^{-1/2}
			\|v\|_{L^2}.
			\label{eq:averaging-code-bound}
		\end{equation}
		Differentiate the fixed factor $\overline{\gamma_S}$ and use
		\eqref{eq:averaging-code-bound}. This gives
		\eqref{eq:averaging-branch-Cr-bound} and
		\eqref{eq:averaging-branch-Hr-bound}. Note that, realification is an isometry and
		doubles the finite-dimensional real range dimension at most.
\end{proof}

\subsection{Auxiliary results for the universality proof}
\label{subsec:auxiliary-metaplectic neural operator-universality}

This subsection develops the auxiliary results used in the proof of the
metaplectic neural operator universal approximation theorem.

\subsubsection{Finite-code averaging operators}
\label{subsec:aux-finite-code-averaging}

The replacement for the Fourier coefficient vector is a nonlinear finite
code obtained by averaging a local feature map.  We first isolate this
normal form and its continuity properties.

\begin{definition}[Finite-code averaging operator]
	\label{def:aux-finite-code-operator}
	Let $m\in\mathbb N$.  Let
	\begin{equation}
		R:\mathbb R^{d_a}\times\mathbb R^d
		\longrightarrow\mathbb R^m,
		\qquad
		Q:\mathbb R^m\times\mathbb R^d
		\longrightarrow\mathbb R^{d_u}
		\label{eq:aux-R-Q-domains}
	\end{equation}
	be finite-dimensional affine-output neural networks with activation
	$\sigma$.  For $a\in L^2(\Omega;\mathbb R^{d_a})$, define
	\begin{equation}
		\mathbf c_R(a)
		:=
		\operatorname{Avg}_\Omega
		\bigl(R(a(\cdot),\cdot)\bigr)
		\in\mathbb R^m
		\label{eq:aux-finite-code}
	\end{equation}
	and
	\begin{equation}
		\mathscr T_{R,Q}(a)(x)
		:=
		Q(\mathbf c_R(a),x),
		\qquad x\in\Omega.
		\label{eq:aux-finite-code-operator}
	\end{equation}
\end{definition}

\begin{lemma}[Continuity and compactness of the finite code]
	\label{lem:aux-code-continuity-compactness}
	Let $R$ and $Q$ be as in Definition
	\ref{def:aux-finite-code-operator}.  There are constants $L_R,C_R\geq0$
	such that
	\begin{equation}
		|R(z,x)-R(z',x)|
		\leq L_R|z-z'|,
		\qquad
		|R(0,x)|\leq C_R
		\label{eq:aux-R-Lipschitz-growth}
	\end{equation}
	for all $z,z'\in\mathbb R^{d_a}$ and $x\in\overline\Omega$.  Consequently,
	$\mathbf c_R:L^2(\Omega;\mathbb R^{d_a})\to\mathbb R^m$ is well defined and
	satisfies
	\begin{align}
		|\mathbf c_R(a)-\mathbf c_R(b)|
		&\leq
		L_R|\Omega|^{-1/2}\|a-b\|_{L^2},
		\label{eq:aux-code-L2-Lipschitz}
		\\
		|\mathbf c_R(a)|
		&\leq
		C_R+L_R|\Omega|^{-1/2}\|a\|_{L^2}.
		\label{eq:aux-code-growth}
	\end{align}
	
The following set     $
		\mathfrak C_R(K)
		:=
		\{\mathbf c_R(a):a\in K\}
		\subset\mathbb R^m
	$
	is compact.
	
	For every $r\in\mathbb N_0$, the map
	$$
		\mathscr T_{R,Q}:
		L^2(\Omega;\mathbb R^{d_a})
		\longrightarrow
		C^r(\overline\Omega;\mathbb R^{d_u})
	$$
	is continuous.
\end{lemma}

\begin{proof}
	A finite neural network built from affine maps and the globally Lipschitz
	activation $\sigma$ is globally Lipschitz.  This gives the first estimate in
	\eqref{eq:aux-R-Lipschitz-growth}; the second follows from continuity of
	$x\mapsto R(0,x)$ on the compact set $\overline\Omega$.  Hence
	$$
		|R(a(x),x)|
		\leq C_R+L_R|a(x)|
	$$
	for almost every $x$, so the average in \eqref{eq:aux-finite-code} is well
	defined.  The estimates \eqref{eq:aux-code-L2-Lipschitz} and
	\eqref{eq:aux-code-growth} follow from Cauchy--Schwarz.  Compactness of
	$\mathfrak C_R(K)$ follows because it is the continuous image of $K$.
	
	Since $\sigma\in C^\infty(\mathbb R)$, the finite-dimensional network $Q$
	is smooth.  For each multi-index $\alpha$, the function
	$$
		(c,x)\longmapsto\partial_x^\alpha Q(c,x)
	$$
	is continuous.  Uniform continuity on compact subsets of
	$\mathbb R^m\times\overline\Omega$ shows that
	$c\mapsto Q(c,\cdot)$ is continuous from $\mathbb R^m$ into $C^r$.  The
	last assertion is the composition of this map with the continuous code map
	$\mathbf c_R$.
\end{proof}

The following proposition records the precise external consequence of the
averaging-neural-operator theorem that is used later.

\begin{proposition}
	\label{prop:aux-ANO-finite-code-normal-form}
	Suppose that   \cref{ass:activation} holds, and let
	$s,r\in\mathbb N_0$.
	Let $\mathcal G:X\to Y$ be continuous and let $K\subset X$ be compact,
	in either of the following settings:
	\begin{enumerate}
		\item[\emph{(i)}] \emph{$C^s$ setting:}
			$X=C^s(\overline\Omega;\mathbb R^{d_a})$
			and 
			$Y=C^r(\overline\Omega;\mathbb R^{d_u})$.
		
		\item[\emph{(ii)}] \emph{Sobolev setting:}
			$X=H^s(\Omega;\mathbb R^{d_a})$
			and 
			$Y=H^r(\Omega;\mathbb R^{d_u})$,
		with the additional assumption
		$$
			\sup_{a\in K}\|a\|_{L^\infty}<\infty.
		$$
	\end{enumerate}
	Then, for every $\varepsilon>0$, there exist $m\in\mathbb N$
	and finite neural networks $R,Q$ as in
	\eqref{eq:aux-R-Q-domains} such that
	$$
		\sup_{a\in K}
		\|\mathcal G(a)-\mathscr T_{R,Q}(a)\|_Y
		<\varepsilon.
	$$
\end{proposition}

\begin{proof}
The corresponding ANO universality results are proved in
	\cite[Theorems~1 and~2]{LanthalerLiStuart2025}.
	The proof of Lemma~A.9 and Remark~A.10 in
	\cite{LanthalerLiStuart2025} gives the more precise normal form above.  Its
	encoder constructs a local neural feature map $R(a(x),x)$ and uses a single average over $\Omega$ to obtain a finite vector.  Because the average is
	normalized, a constant hidden bias may be absorbed into the affine output of
	$R$; the hidden componentwise activation and all subsequent
	finite-dimensional processing may then be composed into the local decoder
	$Q(c,x)$.  Thus the resulting operator has the form
	\eqref{eq:aux-finite-code-operator} and uses one evaluation of the averaging
	operation.  The two norm estimates are the $C^s\to C^{s'}$ theorem and the
	$H^s\to H^{s'}$ specialization of the Sobolev theorem in that
	reference.  
\end{proof}

\subsubsection{Local chirp encoding}

Fix for the remainder of this section a free symplectic matrix $S$, and the chirp $\gamma_S$ from
\eqref{eq:free-matrices-and-chirp}.  We use the positional vector $p_S$ from
\eqref{eq:canonical-metaplectic-position}.  

\begin{lemma}
	\label{lem:aux-local-chirp-encoder}
	Let $M>0$, let $m\in\mathbb N$, and let
	$R:\mathbb R^{d_a}\times\mathbb R^d\to\mathbb R^m$ be a neural
	network with activation $\sigma$.  There exist a vector
	$a_R\in\mathbb R^{2m}$ and a number $\tau_R>0$, independent of the
	accuracy below, such that for every $\eta>0$ there is a 
	activated-output network
	$$
		\Phi_{R,\eta}^{\mathrm{enc}}:
		\mathbb R^{d_a+d+2}
		\longrightarrow
		\mathbb R^{2m}
	$$
	satisfying
	\begin{align}
		\sup_{\substack{|z|\leq M\\x\in\overline\Omega}}
		\Bigl|
		\Phi_{R,\eta}^{\mathrm{enc}}(z,p_S(x))
		-a_R
		-\tau_R\mathfrak r_m
		\bigl(\overline{\gamma_S(x)}R(z,x)\bigr)
		\Bigr|
		<\eta.
		\label{eq:aux-local-chirp-encoder-uniform}
	\end{align}
	Its pointwise realization
	$$
		\mathcal R_{R,\eta}(a)(x)
		:=
		\Phi_{R,\eta}^{\mathrm{enc}}(a(x),p_S(x))
	$$
	is an admissible local lifting map of the form
	\eqref{eq:local-lifting-map}.  If $a\in L^\infty$ and
	$\|a\|_{L^\infty}\leq M$, then, with
	$$
		c_R^0:=\mathfrak r_m^{-1}(a_R)\in\mathbb C^m,
	$$
	one has
	\begin{align*}
		\Bigl\|
		\mathfrak r_m^{-1}\mathcal R_{R,\eta}(a)
		-c_R^0
		-\tau_R\overline{\gamma_S}
		R(a(\cdot),\cdot)
		\Bigr\|_{L^2(\Omega;\mathbb C^m)}
		<|\Omega|^{1/2}\eta.
	\end{align*}
\end{lemma}

\begin{proof}
	Consider
	$$
		\mathfrak B_{M,S}
		:=
		\{
		(z,p_S(x)):
		|z|\leq M,\;
		x\in\overline\Omega
		\}
		\subset
		\mathbb R^{d_a+d+2}.
	$$
	This set is compact, since
	$\{(z,x):|z|\leq M,\ x\in\overline\Omega\}$ is compact and
	$(z,x)\mapsto(z,p_S(x))$ is continuous.

	On $\mathfrak B_{M,S}$, define
	$$
		F_{R,S}(z,p_S(x))
		:=
		\mathfrak r_m
		\bigl(
		\overline{\gamma_S(x)}R(z,x)
		\bigr).
	$$
	This is well defined because the first $d$ components of $p_S(x)$
	are the coordinates of $x$, and it is continuous since $R$ is
	continuous, $\gamma_S$ is smooth, and $\mathfrak r_m$ is linear.

	Applying \cref{lem:activated-output-range-encoding} to
	$\mathfrak B_{M,S}$ and $F_{R,S}$ gives fixed
	$a_R\in\mathbb R^{2m}$ and $\tau_R>0$, independent of $\eta$, and,
	for every $\eta>0$, an activated-output network
	$\Phi_{R,\eta}^{\mathrm{enc}}$ satisfying
	\eqref{eq:aux-local-chirp-encoder-uniform}.

	Define
	$$
		\mathcal R_{R,\eta}(a)(x)
		:=
		\Phi_{R,\eta}^{\mathrm{enc}}
		\bigl(a(x),p_S(x)\bigr).
	$$
	This is a local lifting map of the form
	\eqref{eq:local-lifting-map}.

	For $a$ as in the statement, evaluate
	\eqref{eq:aux-local-chirp-encoder-uniform} at $z=a(x)$.  Since
	$\mathfrak r_m$ is an isometry and
	$$
		c_R^0
		:=
		\mathfrak r_m^{-1}(a_R),
	$$
	we obtain, for almost every $x\in\Omega$,
	$$
		\left|
		\mathfrak r_m^{-1}\mathcal R_{R,\eta}(a)(x)
		-
		c_R^0
		-
		\tau_R\overline{\gamma_S(x)}R(a(x),x)
		\right|
		<
		\eta.
	$$
	Integrating this pointwise estimate yields
	$$
		\left\|
		\mathfrak r_m^{-1}\mathcal R_{R,\eta}(a)
		-
		c_R^0
		-
		\tau_R\overline{\gamma_S}
		R(a(\cdot),\cdot)
		\right\|_{L^2}
		<
		|\Omega|^{1/2}\eta,
	$$
	as claimed.
\end{proof}

\subsubsection{One metaplectic branch produces the finite code}
\label{subsec:aux-metaplectic-code-field}

We first record explicitly that the constant vector $c_R^0$ in
Lemma~\ref{lem:aux-local-chirp-encoder} can be canceled by the declared
finite-dimensional bias class.

\begin{lemma}
	\label{lem:aux-chirped-constant-bias}
	Let $d_0=u_0+iv_0\in\mathbb C^m$ and write
	$$
		\varphi_S(x):=\pi x\cdot Q_Sx.
	$$
	Then
	\begin{align}
		\mathfrak r_m
		\bigl(\overline{\gamma_S(x)}d_0\bigr)
		={}&
		\begin{pmatrix}u_0\\v_0\end{pmatrix}
		\cos\varphi_S(x)
		+
		\begin{pmatrix}v_0\\-u_0\end{pmatrix}
		\sin\varphi_S(x).
		\label{eq:aux-chirped-constant-realification}
	\end{align}
	Consequently, the map
	$x\mapsto\mathfrak r_m(\overline{\gamma_S(x)}d_0)$ belongs to the bias
	class in Definition~\ref{def:finite-parameter-spatial-bias} with positional
	vector $p_S$.
\end{lemma}

\begin{proof}
	The identity follows by multiplying $u_0+iv_0$ by
	$e^{-i\varphi_S(x)}$.  The last two entries of $p_S$ are
	$\cos\varphi_S$ and $\sin\varphi_S$, so the right-hand side of
	\eqref{eq:aux-chirped-constant-realification} is obtained by a fixed
	$2m\times(d+2)$ coefficient matrix whose first $d$ columns are zero.
\end{proof}

\begin{lemma}[Uniform activation interval for a compact code set]
	\label{lem:aux-code-activation-window}
	Let $I_\sigma$ be the interval from \cref{lem:local-inverse-activation} and
    $C\subset\mathbb R^m$ be compact.  There exist compact intervals
	$$
		I_0\Subset I_1\Subset I_\sigma,
		\qquad
		t_*\in\operatorname{int}(I_0),
	$$
	and a number $\lambda>0$ such that
	\begin{equation}
		t_*\mathbf 1_{2m}
		+\lambda\mathfrak r_m
		\bigl(\overline{\gamma_S(x)}c\bigr)
		\in I_0^{2m}
		\label{eq:aux-exact-preactivation-window}
	\end{equation}
	for every $c\in C$ and every $x\in\overline\Omega$.  If
	$$
		J_j:=\sigma(I_j),
		\qquad j\in\{0,1\},
	$$
	then $J_0\Subset J_1\Subset J_\sigma$ and the inverse
	$\varsigma:=(\sigma|_{I_\sigma})^{-1}$ is smooth on an open neighborhood of
	$J_1$.
\end{lemma}

\begin{proof}
	Choose compact intervals $I_0\Subset I_1\Subset I_\sigma$ containing $t_*$
	in their interiors.  Since $C$ is compact, there exists $M_C$ with
	$|c|\leq M_C$ for $c\in C$.  Every coordinate of
	$\mathfrak r_m(\overline{\gamma_S(x)}c)$ has absolute value at most $M_C$.
	Taking $\lambda>0$ sufficiently small gives
	\eqref{eq:aux-exact-preactivation-window}.  Since
	$\sigma|_{I_\sigma}$ is a diffeomorphism, it maps compactly contained
	intervals to compactly contained intervals, and its inverse is smooth on
	$J_\sigma$.
\end{proof}

We will later compare two functions
$u,v\in C^r(\overline\Omega;\mathbb R^n)$
before applying the activation.  We therefore need to control the
difference
$\Sigma_n(u)-\Sigma_n(v)$
in the $C^r$ norm in terms of $u-v$.

We record the corresponding estimate in the following lemma.

\begin{lemma}
	\label{lem:aux-Cr-activation-stability}
	Let $r,n\in\mathbb N_0$ with $n\geq1$, let $I\subset\mathbb R$ be compact,
	and let $M>0$.  There exists a constant
	$C=C(r,n,I,M,\sigma)$ such that
	\begin{equation*}
		\|\Sigma_n(u)-\Sigma_n(v)\|_{C^r}
		\leq
		C\|u-v\|_{C^r}
	\end{equation*}
	whenever $u,v\in C^r(\overline\Omega;\mathbb R^n)$ satisfy
	$$
		u(x),v(x)\in I^n,
		\qquad
		\|u\|_{C^r}+\|v\|_{C^r}\leq M.
	$$
\end{lemma}
\begin{proof}
Write $u=(u_1,\ldots,u_n)$ and $v=(v_1,\ldots,v_n)$.
Since $\Sigma_n$ acts componentwise, global Lipschitz continuity
of $\sigma$ gives
$$
\|\Sigma_n(u)-\Sigma_n(v)\|_{C^0}
\le L_\sigma\|u-v\|_{C^0},
$$
which proves the case $r=0$.

Let $r\ge1$ and fix a component $\ell$.
For $1\le|\alpha|\le r$, the chain rule expresses
$\partial^\alpha(\sigma\circ u_\ell)$ as a finite linear
combination of terms
$$
\sigma^{(k)}(u_\ell)
\prod_{j=1}^k\partial^{\beta_j}u_\ell,
\qquad
|\beta_j|\ge1,
\qquad
\sum_{j=1}^k\beta_j=\alpha.
$$
Compare each term with its counterpart for $v_\ell$.

By changing one factor at a time, using
$ab-cd=(a-c)b+c(b-d)$, their difference becomes a sum of terms
containing either $\partial^\beta(u_\ell-v_\ell)$ or
$\sigma^{(k)}(u_\ell)-\sigma^{(k)}(v_\ell)$.

Let $\widetilde I$ be the smallest closed interval containing $I$.
The mean-value theorem gives
$$
|\sigma^{(k)}(u_\ell)-\sigma^{(k)}(v_\ell)|
\le
\sup_{t\in\widetilde I}|\sigma^{(k+1)}(t)|
\,|u_\ell-v_\ell|.
$$
Moreover,
$$
|\partial^\beta(u_\ell-v_\ell)|
\le \|u_\ell-v_\ell\|_{C^r},
\qquad |\beta|\le r.
$$
All remaining factors are uniformly bounded by the $C^r$ bound
$M$ and the bounds on the derivatives of $\sigma$ on $\widetilde I$.
Summing the terms yields
$$
\|\sigma\circ u_\ell-\sigma\circ v_\ell\|_{C^r}
\le C_0\|u_\ell-v_\ell\|_{C^r},
\qquad C_0=C_0(r,I,M,\sigma).
$$
Combining the $n$ component estimates gives
$$
\|\Sigma_n(u)-\Sigma_n(v)\|_{C^r}
\le \sqrt n\,C_0\|u-v\|_{C^r}.
$$
Thus the lemma  holds with $C=\sqrt n\,C_0$.
\end{proof}

\subsubsection{Local demodulation and decoding}
\label{subsec:aux-local-demodulation-decoding}
Later, the output will be obtained by applying a finite-dimensional map
$F$ pointwise to a function $h$, namely,
$$
	x\longmapsto F(h(x),x).
$$
We need two facts about this substitution.  First, if $h$ is replaced by
a nearby function $\widetilde h$, then the corresponding output should
remain close in the $C^r$ norm.  Second, if $F$ is approximated by
another  map $G$, then this approximation should
remain valid after substituting $h(x)$.  The next lemma provides both
estimates uniformly for $h$ near a compact family of functions.

\begin{lemma}
	\label{lem:aux-Cr-substitution}
	Let $r\in\mathbb N_0$ and $d,n,p\in\mathbb N$.  Let
	$U\subset\mathbb R^n$ and $V\subset\mathbb R^d$ be open with
	$\overline\Omega\subset V$.  Let
	$H\subset C^r(\overline\Omega;\mathbb R^n)$ be compact and suppose that
	$$
		\{h(x):h\in H,\ x\in\overline\Omega\}
		\Subset U.
	$$
	If $F\in C^{r+1}(U\times V;\mathbb R^p)$, then there are a bounded
	open neighborhood $\mathcal U_H$ of $H$ in $C^r$ and a constant
	$C_F>0$ such that
	\begin{equation}
		\|F(h(\cdot),\cdot)-F(\widetilde h(\cdot),\cdot)\|_{C^r}
		\leq
		C_F\|h-\widetilde h\|_{C^r}
		\label{eq:aux-substitution-local-Lipschitz}
	\end{equation}
	for all $h,\widetilde h\in\mathcal U_H$.

	The neighborhood $\mathcal U_H$ can be chosen so that there exists a
	compact set $K_0\Subset U\times V$ containing
	\begin{equation}
		\{(h(x),x):h\in\mathcal U_H,\ x\in\overline\Omega\}.
		\label{eq:aux-substitution-composition-compact}
	\end{equation}
	For this choice of $\mathcal U_H$ and $K_0$, there exists a constant
	$C_H>0$ such that, for every
	$G\in C^r(U\times V;\mathbb R^p)$,
	\begin{equation}
		\sup_{h\in\mathcal U_H}
		\|F(h(\cdot),\cdot)-G(h(\cdot),\cdot)\|_{C^r}
		\leq
		C_H\|F-G\|_{C^r(K_0)}.
		\label{eq:aux-substitution-function-approximation}
	\end{equation}
\end{lemma}
\begin{proof}
The set
$$
 K_H:=\{h(x):h\in H,\ x\in\overline\Omega\}
$$
is compact by continuity of evaluation. Choose $\rho>0$ such that
$$
 K_U:=\{z:\operatorname{dist}(z,K_H)\le\rho\}\Subset U,
$$
and define
$$
 \mathcal U_H
 :=\{h\in C^r:\operatorname{dist}_{C^r}(h,H)<\rho/4\}.
$$
Then $\mathcal U_H$ is a bounded open neighborhood of $H$,
and its functions take values in $K_U$. Consequently,
$$
 K_0:=K_U\times\overline\Omega\Subset U\times V
$$
satisfies \eqref{eq:aux-substitution-composition-compact}, and
$\sup_{h\in\mathcal U_H}\|h\|_{C^r}<\infty$.

 Now want to show the stability under perturbations of $h$.
 
Suppose first that $\|h-\widetilde h\|_{C^r}<\rho/2$.
For $h_t:=(1-t)h+t\widetilde h$, we have
$\operatorname{dist}(h_t(x),K_H)<3\rho/4$, hence
$h_t(x)\in K_U$. The fundamental theorem of calculus gives
$$
 F(h(x),x)-F(\widetilde h(x),x)
 =
 \int_0^1 D_zF(h_t(x),x)
 \bigl(h(x)-\widetilde h(x)\bigr)\,dt.
$$
Since the $h_t$ are uniformly bounded in $C^r$ and
$F\in C^{r+1}$, the chain and product rules bound the
$C^r$ norm of this expression by
$C_1\|h-\widetilde h\|_{C^r}$.

The chain rule also gives a uniform bound
$\|F(h(\cdot),\cdot)\|_{C^r}\le B$ on $\mathcal U_H$.
For the remaining pairs,
$$
 \|F(h(\cdot),\cdot)-F(\widetilde h(\cdot),\cdot)\|_{C^r}
 \le 2B
 \le \frac{4B}{\rho}\|h-\widetilde h\|_{C^r}.
$$
Thus \eqref{eq:aux-substitution-local-Lipschitz} holds with
$C_F:=\max\{C_1,4B/\rho\}$.

\medskip
Now we  want an approximation of the pointwise map.
Apply the chain rule to $(F-G)(h(x),x)$.
Each term contains a derivative of $F-G$ of order at most $r$,
evaluated in $K_0$, multiplied by derivatives of $h$.
The uniform $C^r$ bound on $\mathcal U_H$ therefore yields
$$
 \sup_{h\in\mathcal U_H}
 \|F(h(\cdot),\cdot)-G(h(\cdot),\cdot)\|_{C^r}
 \le C_H\|F-G\|_{C^r(K_0)},
$$
where $C_H$ is independent of $G$. This proves
\eqref{eq:aux-substitution-function-approximation}.
\end{proof}

We conclude by recording the two fixed linear maps used to recover a real
averaged vector from its chirp-modulated representation.  Write
$\gamma_S(x)=e^{i\varphi_S(x)}$.  Multiplication by $\gamma_S(x)$ on
$\mathbb C^m$ is represented, after realification, by the real rotation
$$
	\mathscr U_S(x)
	:=
	\begin{pmatrix}
		\cos\varphi_S(x)I_m
		&
		-\sin\varphi_S(x)I_m
		\\
		\sin\varphi_S(x)I_m
		&
		\cos\varphi_S(x)I_m
	\end{pmatrix}
	\in\mathbb R^{2m\times2m}.
$$
Thus, for every $z\in\mathbb C^m$,
\begin{equation*}
	\mathscr U_S(x)\mathfrak r_m(z)
	=
	\mathfrak r_m\bigl(\gamma_S(x)z\bigr).
\end{equation*}

We also define the fixed real-linear map
$$
	\Pi_{\mathrm{re}}:
	\mathbb R^{2m}
	\longrightarrow
	\mathbb R^m,
	\qquad
	\Pi_{\mathrm{re}}
	\begin{pmatrix}
		u\\v
	\end{pmatrix}
	:=
	u.
$$
The map $\Pi_{\mathrm{re}}$ is applied only after the chirp modulation
has been removed.  Indeed, if $c\in\mathbb R^m$, then
\begin{equation*}
\Pi_{\mathrm{re}}\mathscr U_S(x)\mathfrak r_m\bigl(\overline{\gamma_S(x)}c\bigr)=\Pi_{\mathrm{re}}\mathfrak r_m(c)=c.
\end{equation*} 

Thus, the rotation removes the factor
$\overline{\gamma_S(x)}$, after which $\Pi_{\mathrm{re}}$ recovers the
original real vector.  Both maps are fixed and introduce no trainable
parameters. 

\section{Universal approximation by metaplectic neural operators}
\label{sec:universal-approximation-metaplectic neural operator}

We formulate the main approximation theorem in two function-space settings.
We retain the nonempty bounded Lipschitz domain $\Omega\subset\mathbb R^d$
and the channel dimensions $d_a,d_u\in\mathbb N$ fixed in
\cref{sec:preliminaries-direct-metaplectic neural operator}.
The integers $s,r\in\mathbb N_0$ denote the input and output regularity orders.
For a continuous operator $\mathcal G:X\to Y$ and a compact set $K\subset X$,
we consider the following settings:
\begin{enumerate}
    \item[\emph{(i)}] \emph{$C^s$ setting:}
    $X=C^s(\overline\Omega;\mathbb R^{d_a})$ and
    $Y=C^r(\overline\Omega;\mathbb R^{d_u})$.

    \item[\emph{(ii)}] \emph{Sobolev setting:}
    $X=H^s(\Omega;\mathbb R^{d_a})$ and
    $Y=H^r(\Omega;\mathbb R^{d_u})$, with the additional assumption
    $$
        \sup_{a\in K}\|a\|_{L^\infty(\Omega)}<\infty.
    $$
\end{enumerate}
In setting \emph{(i)}, compact subsets of $X$ are automatically bounded in
$L^\infty$. In setting \emph{(ii)}, this bound holds automatically when
$s>d/2$, by the Sobolev embedding on Lipschitz
domains~\cite{AdamsFournier2003}. For $s\le d/2$, it remains an additional
assumption, which also appears in the Sobolev theorem for averaging neural
operators~\cite{LanthalerLiStuart2025}.

To handle both output norms together, for each $\varepsilon>0$ we set
\begin{equation}\label{eq:deltaError}
    C_Y:=
    \begin{cases}
        1, & Y=C^r,\\
        |\Omega|^{1/2}\binom{d+r}{r}^{1/2}, & Y=H^r,
    \end{cases}
    \qquad
    \delta:=\varepsilon/C_Y.
\end{equation}
Then, for every $w\in C^r(\overline\Omega;\mathbb R^{d_u})$,
$$
    \|w\|_Y\le C_Y\|w\|_{C^r}.
$$

The theorem holds for every fixed free symplectic matrix $S$. Its
approximants are built from the transform $\widehat S$, the atom
$\rho_{\widehat S,\Omega}$ in \eqref{eq:metaplectic-averaging-symbol} and the
positional vector $p_S$ in \eqref{eq:canonical-metaplectic-position}.
These are fixed before $\mathcal G$, $K$ and the accuracy $\varepsilon$
are chosen, whereas the number of channels, the coefficients and the
pointwise networks may depend on these data.

\begin{theorem}
	\label{thm:universal-approximation-metaplectic neural operator}
	Suppose that \cref{ass:activation} holds, and fix a free
	symplectic matrix $S$.
    
	Let $\mathcal G:X\to Y$ be continuous and let $K\subset X$
	be compact, with $X$, $Y$, and $K$ satisfying the assumptions
	of either setting \emph{(i)} or \emph{(ii)} above.
	Then, for every $\varepsilon>0$, there exists a metaplectic
	neural operator $\mathcal N$ such that
	$$
		\sup_{a\in K}
		\|\mathcal G(a)-\mathcal N(a)\|_Y
		<\varepsilon.
	$$
\end{theorem}

\begin{proof}
Let $x\in\Omega$. Fix $\varepsilon>0$, and let $C_Y$ and $\delta$ be as in \eqref{eq:deltaError}.  By proposition~\ref{prop:aux-ANO-finite-code-normal-form}, there exist $m\in\mathbb N$ and maps $R$ and $Q$, with the domains specified in
\eqref{eq:aux-R-Q-domains}, such that the finite averaged-code operator
$$
 \mathscr T_{R,Q}(a)(x)=Q(\mathbf c_R(a),x),
$$
Such that   $\mathbf c_R(a)=\operatorname{Avg}_\Omega R(a(\cdot),\cdot)$,
defined in \eqref{eq:aux-finite-code-operator}, satisfies
\begin{equation}\label{eq:main-Part-I-ANO-error}
 \sup_{a\in K}
 \|\mathcal G(a)-\mathscr T_{R,Q}(a)\|_Y
 <\frac{\varepsilon}{2}.
\end{equation}
We now construct a metaplectic neural operator that approximates
$\mathscr T_{R,Q}$ on $K$. 
More precisely, we will show that
\begin{equation}\label{eq:main-Part-II-goal}
 \sup_{a\in K}
 \|\mathscr T_{R,Q}(a)-\mathcal N(a)\|_{C^r}
 <\frac{\delta}{2}.
\end{equation}
For $a\in K$, introduce the following function 
\begin{equation*}
 v_a(y):=\overline{\gamma_S(y)}R(a(y),y),
 \qquad y\in\Omega.
\end{equation*}
The uniform $L^\infty$ bound on $K$ and the continuity of $R$ ensure that $v_a\in L^2(\Omega;\mathbb C^m)$. 
By   Proposition~\ref{prop:exact-chirped-average} we get
\begin{equation}\label{eq:main-Part-II-exact-code-branch}
 \bigl(\mathcal K_{\widehat S,\rho_{\widehat S,\Omega}I_m}^{\Omega}
 v_a\bigr)(x)
 =\overline{\gamma_S(x)}\mathbf c_R(a).
\end{equation}
To approximate $v_a$ by an admissible local lifting we  apply \cref{lem:aux-local-chirp-encoder}. which  provides constants $c_R^0\in\mathbb C^m$ and $\tau_R>0$ and, for every $\eta>0$, an activated-output neural network $\Phi^{\mathrm{enc}}_{R,\eta}$ such that
$$
 \mathcal R_{R,\eta}(a)(x)
 :=\Phi^{\mathrm{enc}}_{R,\eta}\bigl(a(x),p_S(x)\bigr)
 \in\mathbb R^{2m}.
$$
 Here $p_S$ is the positional vector from \eqref{eq:canonical-metaplectic-position}, and $c_R^0$ and $\tau_R$ are independent of $\eta$. Set
\begin{equation}\label{eq:eror_valued_function}
 e_{R,\eta}(a)
 :=\mathfrak r_m^{-1}\mathcal R_{R,\eta}(a)
   -c_R^0-\tau_Rv_a,
\end{equation}
where $\mathfrak r_m$ is  realification function.  Then, \cref{lem:aux-local-chirp-encoder} gives  the uniform estimate
\begin{equation}\label{eq:lifting-error}
 \sup_{a\in K}\|e_{R,\eta}(a)\|_{L^2}
 <|\Omega|^{1/2}\eta.
\end{equation}
The parameter  $\eta$ will be chosen later.

By the compactness of $\mathscr C_{R,K}$ and Lemma~\ref{lem:aux-code-activation-window}, we may fix intervals
$I_0\Subset I_1\Subset I_\sigma$, a point $t_*\in\operatorname{int}(I_0)$, and a number $\lambda>0$ such that
\begin{equation}\label{eq:main-Part-II-auxiliary-window}
 t_*\mathbf1_{2m}
 +\lambda\mathfrak r_m\bigl(\overline{\gamma_S(x)}c\bigr)
 \in I_0^{2m}
 \qquad
 (c\in\mathscr C_{R,K},\ x\in\overline\Omega).
\end{equation}
Set $\alpha:=\lambda/\tau_R$ and choose the  matrix-valued symbol  
$$
 P_1(\xi):=\alpha\rho_{\widehat S,\Omega}(\xi)I_m,
 \qquad \xi\in\mathbb R^d.
$$
Identifying $c_R^0$ with the corresponding constant function on $\Omega$, linearity and \eqref{eq:eror_valued_function} give
$$
 \begin{aligned}
 \mathcal K_{\widehat S,P_1}^{\Omega}
 \bigl(\mathfrak r_m^{-1}\mathcal R_{R,\eta}(a)\bigr)
 &={}
 \mathcal K_{\widehat S,P_1}^{\Omega}c_R^0
 +\tau_R\mathcal K_{\widehat S,P_1}^{\Omega}v_a
 +\mathcal K_{\widehat S,P_1}^{\Omega}e_{R,\eta}(a).
 \end{aligned}
$$
Proposition~\ref{prop:exact-chirped-average} and \eqref{eq:main-Part-II-exact-code-branch} identify the first two terms explicitly:
$$
 \begin{aligned}
 \bigl(\mathcal K_{\widehat S,P_1}^{\Omega}c_R^0\bigr)(x)
 &=\overline{\gamma_S(x)}d_R^0, 
 \\
 \tau_R\bigl(\mathcal K_{\widehat S,P_1}^{\Omega}v_a\bigr)(x)
 &=\lambda\overline{\gamma_S(x)}\mathbf c_R(a),
 \end{aligned}
$$
where $ d_R^0:=\alpha c_R^0\operatorname{Avg}_\Omega(\gamma_S).$ The first term is independent of $a$, so it can be canceled by the
spatial bias
$$
 b_1(x)
 :=t_*\mathbf1_{2m}
   -\mathfrak r_m\bigl(\overline{\gamma_S(x)}d_R^0\bigr).
$$
By Lemma~\ref{lem:aux-chirped-constant-bias}, this bias belongs to the
class in \cref{def:finite-parameter-spatial-bias}. In particular, there exist $c_1\in\mathbb R^{2m}$ and $G_1\in\mathbb R^{2m\times(d+2)}$ such that
$$
 b_1(x)=c_1+G_1p_S(x).
$$
Taking $W_1=0$, define the admissible hidden layer
\begin{equation*}
 \mathcal L_1(w)(x)
 :=\Sigma_{2m}\!\left(
 b_1(x)+\bigl(\mathbf K_{\widehat S,P_1}^{\Omega}w\bigr)(x)
 \right).
\end{equation*} 
The resulting preactivation is given by 
\begin{equation}\label{eq:main-actual-preactivation}
 u_{R,\eta}(a)(x)
 :=
 b_1(x)
 +\bigl(
 \mathbf K_{\widehat S,P_1}^{\Omega}
 \mathcal R_{R,\eta}(a)
 \bigr)(x).
\end{equation}
Also let 
\begin{equation}\label{eq:main-exact-code-preactivation}
 u_R(a)(x)
 :=
 t_*\mathbf1_{2m}
 +\lambda\mathfrak r_m\bigl(
 \overline{\gamma_S(x)}\mathbf c_R(a)
 \bigr).
\end{equation}
Then following  estimate holds: 
$$
 \sup_{a\in K}
 \|u_{R,\eta}(a)-u_R(a)\|_{C^r}
 <C_{\mathrm{pre}}\eta,
$$
where $C_{\mathrm{pre}}>0$ is independent of $a$ and $\eta$. 
Moreover for $\eta$  sufficiently small,  we get  $u_{R,\eta}(a)$ takes values in $I_1^{2m}$. 

Indeed, using the definition of $b_1$, we obtain
$$
\begin{aligned}
u_{R,\eta}(a)(x)
&= b_1(x)
   +\bigl(\mathbf K^\Omega_{\widehat S,P_1}
     \mathcal R_{R,\eta}(a)\bigr)(x)
\\
&= b_1(x)
   +\mathfrak r_m\!\left(
     \Bigl(
       \mathcal K^\Omega_{\widehat S,P_1}
       \bigl(\mathfrak r_m^{-1}\mathcal R_{R,\eta}(a)\bigr)
     \Bigr)(x)
   \right)
\\
&= t_\ast\mathbf1_{2m}
   -\mathfrak r_m\!\left(
     \overline{\gamma_S(x)}\,d_R^0
   \right)
\\
&\quad
   +\mathfrak r_m\!\left(
     \overline{\gamma_S(x)}\,d_R^0
     +\lambda\,\overline{\gamma_S(x)}\,\mathbf c_R(a)
     +\bigl(
       \mathcal K^\Omega_{\widehat S,P_1}e_{R,\eta}(a)
     \bigr)(x)
   \right)
\\
&= t_\ast\mathbf1_{2m}
   +\underbrace{
     -\mathfrak r_m\!\left(
       \overline{\gamma_S(x)}\,d_R^0
     \right)
     +\mathfrak r_m\!\left(
       \overline{\gamma_S(x)}\,d_R^0
     \right)
   }_{=\,0}
\\
&\quad
   +\lambda\,\mathfrak r_m\!\left(
     \overline{\gamma_S(x)}\,\mathbf c_R(a)
   \right)
   +\mathfrak r_m\!\left(
     \bigl(
       \mathcal K^\Omega_{\widehat S,P_1}e_{R,\eta}(a)
     \bigr)(x)
   \right)
\\
&= t_\ast\mathbf1_{2m}
   +\lambda\,\mathfrak r_m\!\left(
     \overline{\gamma_S(x)}\,\mathbf c_R(a)
   \right)
   +\mathfrak r_m\!\left(
     \bigl(
       \mathcal K^\Omega_{\widehat S,P_1}e_{R,\eta}(a)
     \bigr)(x)
   \right).
\end{aligned}
$$ 
Then
$$
u_{R,\eta}(a)-u_R(a)
= \mathfrak{r}_m
  \bigl(\mathcal{K}^{\Omega}_{\widehat{S},P_1}
  e_{R,\eta}(a)\bigr).
$$
Since realification preserves the $C^r$ norm, (36) and (61) give
$$
\begin{aligned}
\sup_{a\in K}
\bigl\|u_{R,\eta}(a)-u_R(a)\bigr\|_{C^r}
&= \sup_{a\in K}
   \bigl\|\mathcal{K}^{\Omega}_{\widehat{S},P_1}
   e_{R,\eta}(a)\bigr\|_{C^r} \\
&\le C^{\mathbb{C}}_{r,S,\Omega}\,
   \alpha |\Omega|^{-1/2}
   \sup_{a\in K}\bigl\|e_{R,\eta}(a)\bigr\|_{L^2} \\
&< C^{\mathbb{C}}_{r,S,\Omega}
   \frac{\lambda}{\tau_R}\,\eta.
\end{aligned}
$$ 
Since realification preserves the $C^r$ norm, \eqref{eq:averaging-branch-Cr-bound} and \eqref{eq:lifting-error} therefore imply
$$
 \begin{aligned}
 \sup_{a\in K}\|u_{R,\eta}(a)-u_R(a)\|_{C^r}
 &\le C^{C}_{r,S,\Omega}\alpha|\Omega|^{-1/2}
       \sup_{a\in K}\|e_{R,\eta}(a)\|_{L^2}
 \\
 &<C^{C}_{r,S,\Omega}\frac{\lambda}{\tau_R}\eta.
 \end{aligned}
$$
Hence, with $C_{\mathrm{pre}}:=C^{C}_{r,S,\Omega}\lambda/\tau_R$, we have
\begin{equation}\label{eq:main-Part-II-preactivation-error-short}
 \sup_{a\in K}\|u_{R,\eta}(a)-u_R(a)\|_{C^r}
 <C_{\mathrm{pre}}\eta.
\end{equation}
By \eqref{eq:main-Part-II-auxiliary-window}, the  preactivations in \eqref{eq:main-exact-code-preactivation} take values in $I_0^{2m}$. Let
$$
 \delta_I:=\operatorname{dist}(I_0,\mathbb R\setminus I_1)>0
$$
and choose $\eta_0>0$ so that $C_{\mathrm{pre}}\eta_0<\delta_I$. Since$$
\sup_{a\in K}
\bigl\|u_{R,\eta}(a)-u_R(a)\bigr\|_{C^0}
\le
\sup_{a\in K}
\bigl\|u_{R,\eta}(a)-u_R(a)\bigr\|_{C^r}.$$
Then \eqref{eq:main-Part-II-preactivation-error-short} ensures that the preactivations in \eqref{eq:main-actual-preactivation} take values in $I_1^{2m}$ whenever $0<\eta\le\eta_0$.

Both preactivation families are also uniformly bounded in $C^r$. Since, the compactness of $\mathscr C_{R,K}$ and the smoothness of $\gamma_S$ give
$$
 M_0:=\sup_{a\in K}\|u_R(a)\|_{C^r}<\infty,
$$
while \eqref{eq:main-Part-II-preactivation-error-short} yields
$$
 \sup_{a\in K}\|u_{R,\eta}(a)\|_{C^r}
 \le M_0+C_{\mathrm{pre}}\eta_0
 \qquad(0<\eta\le\eta_0).
$$
We may therefore apply Lemma~\ref{lem:aux-Cr-activation-stability} on $I_1$ with a common
$C^r$ bound. Defining
\begin{equation}\label{eq:H1eta-hR}
\mathcal{H}_{1,\eta}(a)
:= \mathcal{L}_1\bigl(\mathcal{R}_{R,\eta}(a)\bigr),
\qquad
h_R(a) := \Sigma_{2m}\bigl(u_R(a)\bigr).
\end{equation} 

we obtain a constant $C_{\mathrm{act}}>0$, independent of $a$ and $0<\eta\le\eta_0$, for which
$$
 \|\mathcal H_{1,\eta}(a)-h_R(a)\|_{C^r}
 \le C_{\mathrm{act}}\|u_{R,\eta}(a)-u_R(a)\|_{C^r}.
$$
In particular,
\begin{equation}\label{eq:main-Part-II-actual-auxiliary-hidden-error}
 \sup_{a\in K}
 \|\mathcal H_{1,\eta}(a)-h_R(a)\|_{C^r}
 <C_{\mathrm{hid}}\eta,
 \qquad
 C_{\mathrm{hid}}:=C_{\mathrm{act}}C_{\mathrm{pre}}.
\end{equation}
Writing $J_j:=\sigma(I_j)$ for $j=0,1$ and $J_\sigma:=\sigma(I_\sigma)$, we also have $ h_R(a)(x)\in J_0^{2m}$ and  $\mathcal H_{1,\eta}(a)(x)\in J_1^{2m}.$
We next construct a smooth pointwise map $ \mathfrak q_{R,Q}$  that recovers $\mathscr T_{R,Q}$  . 
Let $\varsigma:=(\sigma|_{I_\sigma})^{-1}$ and let $\boldsymbol\varsigma_{2m}$ act componentwise:
$$
 \boldsymbol\varsigma_{2m}(h_1,\ldots,h_{2m})
 :=
 \bigl(\varsigma(h_1),\ldots,\varsigma(h_{2m})\bigr),
 \qquad h\in J_\sigma^{2m}.
$$
Since $h_R(a)=\Sigma_{2m}(u_R(a))$ and
$u_R(a)(x)\in I_0^{2m}\subset I_\sigma^{2m}$, we have
$$
 \boldsymbol\varsigma_{2m}(h_R(a)(x))
 =
 \boldsymbol\varsigma_{2m}
 \bigl(\Sigma_{2m}(u_R(a)(x))\bigr)
 =
 u_R(a)(x).
$$
Substituting the definition of $u_R(a)$ gives
$$
 \frac{
 \boldsymbol\varsigma_{2m}(h_R(a)(x))-t_*\mathbf1_{2m}
 }{\lambda}
 =
 \mathfrak r_m
 \bigl(\overline{\gamma_S(x)}\mathbf c_R(a)\bigr).
$$

Recall that
$\mathscr U_S(x)\mathfrak r_m(z)
 =\mathfrak r_m(\gamma_S(x)z)$,  since $|\gamma_S(x)|=1$:
$$\mathscr U_S(x)\frac{\boldsymbol\varsigma_{2m}(h_R(a)(x))-t_*\mathbf1_{2m}}{\lambda}=\mathfrak r_m\bigl(\gamma_S(x)\overline{\gamma_S(x)}\mathbf c_R(a)\bigr)=\mathfrak r_m(\mathbf c_R(a)).$$

Moreover, $\mathbf c_R(a)\in\mathbb R^m$, so its complexification has zero imaginary part. 
Thus, if $\Pi_{\mathrm{re}}:\mathbb R^{2m}\to\mathbb R^m$ denotes projection onto the first $m$ coordinates,
$$
 \mathfrak r_m(\mathbf c_R(a))
 =
 \begin{pmatrix}
 \mathbf c_R(a)\\
 0
 \end{pmatrix},
 \qquad
 \Pi_{\mathrm{re}}\mathfrak r_m(\mathbf c_R(a))
 =
 \mathbf c_R(a).
$$
Consequently,
\begin{equation}\label{eq:main-Part-II-exact-code-recovery}
 \Pi_{\mathrm{re}}\mathscr U_S(x)
 \frac{
 \boldsymbol\varsigma_{2m}(h_R(a)(x))-t_*\mathbf1_{2m}
 }{\lambda}
 =
 \mathbf c_R(a).
\end{equation}

For a bounded open $V\supset\overline\Omega$, define the smooth
auxiliary decoder
\begin{equation*}
 \mathfrak q_{R,Q}(h,x)
 :=
 Q\!\left(
 \Pi_{\mathrm{re}}\mathscr U_S(x)
 \frac{\boldsymbol\varsigma_{2m}(h)-t_*\mathbf1_{2m}}{\lambda},
 x
 \right),
 \qquad
 (h,x)\in J_\sigma^{2m}\times V.
\end{equation*}
Its smoothness follows from that of
$\boldsymbol\varsigma_{2m}$, $\mathscr U_S$, and $Q$.
Substituting \eqref{eq:main-Part-II-exact-code-recovery}, we obtain
the exact identity
\begin{equation}\label{eq:main-Part-II-exact-auxiliary-decoding}
 \mathfrak q_{R,Q}(h_R(a)(x),x)
 =
 Q(\mathbf c_R(a),x)
 =
 \mathscr T_{R,Q}(a)(x).
\end{equation} 
It remains to control the error resulting from replacing $h_R(a)$ by $\mathcal H_{1,\eta}(a)$  from \eqref{eq:H1eta-hR} and to approximate $\mathfrak q_{R,Q}$ by a  neural network.
 Consider
$$
 h_c(x):=\Sigma_{2m}\!\left(
 t_*\mathbf1_{2m}
 +\lambda\mathfrak r_m(\overline{\gamma_S(x)}c)
 \right),
 \qquad c\in\mathscr C_{R,K}.
$$
The map $c\mapsto h_c$ is continuous into $C^r$.
Indeed, for every multi-index $\beta$ with $|\beta|\le r$,
the map
$$
 (c,x)\longmapsto\partial_x^\beta h_c(x)
$$
is continuous on the compact set
$\mathscr C_{R,K}\times\overline\Omega$, hence uniformly
continuous. Therefore, if $c_n\to c$ in $\mathscr C_{R,K}$,
$$
 \sup_{x\in\overline\Omega}
 \left|
 \partial_x^\beta h_{c_n}(x)
 -\partial_x^\beta h_c(x)
 \right|
 \longrightarrow0
 \qquad(|\beta|\le r).
$$
Since only finitely many multi-indices occur, this implies
$\|h_{c_n}-h_c\|_{C^r}\to0$.
Consequently,
$$
 \mathscr H_R
 :=
 \{h_R(a):a\in K\}
 =
 \{h_c:c\in\mathscr C_{R,K}\}
$$
is compact in $C^r(\overline\Omega;\mathbb R^{2m})$.
Moreover, $ \{h(x):h\in\mathscr H_R,\ x\in\overline\Omega\}
 \subset J_0^{2m}\Subset J_\sigma^{2m}.$
Apply \cref{lem:aux-Cr-substitution} to $\mathscr H_R$ and $\mathfrak q_{R,Q}$.
It yields a bounded open neighborhood $\mathcal U_R$ of $\mathscr H_R$, a compact set
$ K_{R,\mathrm{dec}}
 \Subset J_\sigma^{2m}\times V$,
and constants $C_{\mathrm{stab}},C_{\mathrm{comp}}>0$.
In particular, we have
$ \{(h(x),x):h\in\mathcal U_R,\ x\in\overline\Omega\}
 \subset K_{R,\mathrm{dec}}$.
The lemma also gives
\begin{equation}
\label{eq:main-Part-II-decoder-stability}
\|\mathfrak q_{R,Q}(h(\cdot),\cdot)-\mathfrak q_{R,Q}(\widetilde h(\cdot),\cdot)\|_{C^r}
\le C_{\mathrm{stab}}\|h-\widetilde h\|_{C^r},
\end{equation}
and
\begin{equation}\label{eq:composition-estimate}
\begin{aligned}
&\|\mathfrak q_{R,Q}(h(\cdot),\cdot)
  -F(h(\cdot),\cdot)\|_{C^r}
\le C_{\mathrm{comp}}
\|\mathfrak q_{R,Q}-F\|_{C^r(K_{R,\mathrm{dec}})}.
\end{aligned}
\end{equation}
for $h,\widetilde h\in\mathcal U_R$ and smooth $F$.
The second estimate controls all derivatives up to order $r$  after substitution.
We now choose $\eta$ uniformly over $K$.
Since $\mathscr H_R$ is compact and $\mathcal U_R$ is an open
neighborhood of $\mathscr H_R$,
$$
 d_R
 :=
 \operatorname{dist}_{C^r}\!\left(
 \mathscr H_R,\,
 C^r(\overline\Omega;\mathbb R^{2m})\setminus\mathcal U_R
 \right)
 >0.
$$
Choose $0<\eta\le\eta_0$ sufficiently small that
$$
 C_{\mathrm{hid}}\eta
 <
 \min\left\{
 d_R,\frac{\delta}{4C_{\mathrm{stab}}}
 \right\}.
$$
This is possible because all quantities on the right-hand side
are positive and independent of $\eta$.
Fix this $\eta$ and write
$$
 \mathcal H_1
 :=
 \mathcal H_{1,\eta}
 =
 \mathcal L_1\circ\mathcal R_{R,\eta}.
$$
By \eqref{eq:main-Part-II-actual-auxiliary-hidden-error},
$$
 \sup_{a\in K}
 \|\mathcal H_1(a)-h_R(a)\|_{C^r}
 <C_{\mathrm{hid}}\eta<d_R.
$$
Since $h_R(a)\in\mathscr H_R$, the definition of $d_R$ implies $\mathcal H_1(a)\in\mathcal U_R
$ for all $a\in K.$

By \eqref{eq:main-Part-II-exact-auxiliary-decoding},
 $\mathscr T_{R,Q}(a)
 =
 \mathfrak q_{R,Q}(h_R(a)(\cdot),\cdot)$ for all  $a\in K.$
Both $h_R(a)$ and $\mathcal H_1(a)$ belong to $\mathcal U_R$.
Thus, by \cref{eq:main-Part-II-decoder-stability} gives
\begin{equation}\label{eq:main-Part-II-hidden-replacement-error}
\begin{aligned}
\sup_{a\in K}\|\mathscr T_{R,Q}(a)-\mathfrak q_{R,Q}(\mathcal H_1(a)(\cdot),\cdot)\|_{C^r}
&\le C_{\mathrm{stab}}\sup_{a\in K}\|h_R(a)-\mathcal H_1(a)\|_{C^r}\\
&< C_{\mathrm{stab}}C_{\mathrm{hid}}\eta < \frac{\delta}{4}.
\end{aligned}
\end{equation}

It remains to approximate the map $\mathfrak q_{R,Q}$ by a  neural network.  The inclusion $\mathcal H_1(a)\in\mathcal U_R$ ensures that
$$
 \bigl(\mathcal H_1(a)(x),x\bigr)
 \in K_{R,\mathrm{dec}}
 \qquad(a\in K,\ x\in\overline\Omega).
$$
Since $\mathfrak q_{R,Q}$ is smooth on a neighborhood of $K_{R,\mathrm{dec}}$, provides a  affine output 

network
 $\Phi:\mathbb R^{2m+d}\longrightarrow\mathbb R^{d_u}$ such that
$$
 \|\mathfrak q_{R,Q}-\Phi\|_{C^r(K_{R,\mathrm{dec}})}
 <
 \frac{\delta}{4C_{\mathrm{comp}}}.
$$
Define the projection as in \cref{def:finite-direct-metaplectic neural operator} by $$\mathcal Q(h)(x)
 :=
 \Phi(h(x),x).$$ Moreover, the prescribed globally Lipschitz activation gives $\Phi$ at most linear growth. Since $\Omega$ is bounded, this defines a map
$$
 \mathcal Q:
 L^2(\Omega;\mathbb R^{2m})
 \longrightarrow
 L^2(\Omega;\mathbb R^{d_u}).
$$
The network $\Phi$ is smooth, so the composition estimate \eqref{eq:composition-estimate} yields
\begin{equation}\label{eq:main-Part-II-final-network-error}
 \begin{aligned}
 \sup_{a\in K}
 \|\mathfrak q_{R,Q}(\mathcal H_1(a)(\cdot),\cdot)
 -\mathcal Q(\mathcal H_1(a))\|_{C^r}&\le
 \sup_{h\in\mathcal U_R}
 \|\mathfrak q_{R,Q}(h(\cdot),\cdot)
 -\Phi(h(\cdot),\cdot)\|_{C^r}\\
 &\le
 C_{\mathrm{comp}}
 \|\mathfrak q_{R,Q}-\Phi\|_{C^r(K_{R,\mathrm{dec}})}
 <\frac{\delta}{4}.
 \end{aligned}
\end{equation}
Combining
\eqref{eq:main-Part-II-hidden-replacement-error} and
\eqref{eq:main-Part-II-final-network-error}, we obtain
\begin{align*}
\sup_{a\in K}\|\mathscr T_{R,Q}(a)-\mathcal Q(\mathcal H_1(a))\|_{C^r}
&\le \sup_{a\in K}\|\mathscr T_{R,Q}(a)-\mathfrak q_{R,Q}(\mathcal H_1(a)(\cdot),\cdot)\|_{C^r}\\
&\qquad
+ \sup_{a\in K}\|\mathfrak q_{R,Q}(\mathcal H_1(a)(\cdot),\cdot)-\mathcal Q(\mathcal H_1(a))\|_{C^r}\\
&< \frac{\delta}{4}+\frac{\delta}{4}=\frac{\delta}{2}.
\end{align*}

Set $\mathcal N
 :=
 \mathcal Q\circ\mathcal L_1\circ\mathcal R_{R,\eta}$. Since $\mathcal H_1=\mathcal L_1\circ\mathcal R_{R,\eta}$,
the preceding estimate becomes
$$
 \sup_{a\in K}
 \|\mathscr T_{R,Q}(a)-\mathcal N(a)\|_{C^r}
 <\frac{\delta}{2},
$$
which prove \eqref{eq:main-Part-II-goal}.
Finally, the bound $\|w\|_Y\le C_Y\|w\|_{C^r}$ and the identity
$\delta=\varepsilon/C_Y$ imply
$$
 \sup_{a\in K}
 \|\mathscr T_{R,Q}(a)-\mathcal N(a)\|_Y
 \le C_Y\sup_{a\in K}
       \|\mathscr T_{R,Q}(a)-\mathcal N(a)\|_{C^r}
 <\frac{\varepsilon}{2}.
$$
Together with \eqref{eq:main-Part-I-ANO-error}, the triangle inequality
therefore yields
$$
 \begin{aligned}
 \sup_{a\in K}\|\mathcal G(a)-\mathcal N(a)\|_Y
 &\le\sup_{a\in K}\|\mathcal G(a)-\mathscr T_{R,Q}(a)\|_Y
 +\sup_{a\in K}\|\mathscr T_{R,Q}(a)-\mathcal N(a)\|_Y
 \\
 &<\frac{\varepsilon}{2}+\frac{\varepsilon}{2}
 =\varepsilon.
 \end{aligned}
$$

\end{proof}

\section*{Acknowledgments}
Both authors acknowledge the support of the Austrian Science Fund (FWF) through project PAT4788625 (Grant-DOI: 10.55776/PAT4788625).	

\printbibliography

@misc{LiEtAl2020GNO,
  author = {Li, Zongyi and Kovachki, Nikola and Azizzadenesheli, Kamyar and Liu, Burigede and Bhattacharya, Kaushik and Stuart, Andrew and Anandkumar, Anima},
  title = {Neural Operator: Graph Kernel Network for Partial Differential Equations},
  year = {2020},
  eprint = {2003.03485},
  archivePrefix = {arXiv}
}

@article{KovachkiEtAl2023NeuralOperator,
  author = {Kovachki, Nikola and Li, Zongyi and Liu, Burigede and Azizzadenesheli, Kamyar and Bhattacharya, Kaushik and Stuart, Andrew and Anandkumar, Anima},
  title = {Neural Operator: Learning Maps Between Function Spaces With Applications to {PDEs}},
  journal = {Journal of Machine Learning Research},
  year = {2023},
  volume = {24},
  number = {89},
  pages = {1--97},
  url = {https://jmlr.org/papers/v24/21-1524.html}
}

@inproceedings{LiEtAl2020MGNO,
  author = {Li, Zongyi and Kovachki, Nikola and Azizzadenesheli, Kamyar and Liu, Burigede and Stuart, Andrew and Bhattacharya, Kaushik and Anandkumar, Anima},
  title = {Multipole Graph Neural Operator for Parametric Partial Differential Equations},
  booktitle = {Advances in Neural Information Processing Systems},
  volume = {33},
  pages = {6755--6766},
  year = {2020},
  eprint = {2006.09535},
  archivePrefix = {arXiv}
}

@article{ChenChen1995,
  author = {Chen, Tianping and Chen, Hong},
  title = {Universal Approximation to Nonlinear Operators by Neural Networks with Arbitrary Activation Functions and Its Application to Dynamical Systems},
  journal = {IEEE Transactions on Neural Networks},
  year = {1995},
  volume = {6},
  number = {4},
  pages = {911--917},
  doi = {10.1109/72.392253}
}

@article{LuEtAl2021DeepONet,
  author = {Lu, Lu and Jin, Pengzhan and Pang, Guofei and Zhang, Zhongqiang and Karniadakis, George Em},
  title = {Learning Nonlinear Operators via {DeepONet} Based on the Universal Approximation Theorem of Operators},
  journal = {Nature Machine Intelligence},
  year = {2021},
  volume = {3},
  number = {3},
  pages = {218--229},
  doi = {10.1038/s42256-021-00302-5}
}

@article{BhattacharyaEtAl2021PCA,
  author = {Bhattacharya, Kaushik and Hosseini, Bamdad and Kovachki, Nikola B. and Stuart, Andrew M.},
  title = {Model Reduction and Neural Networks for Parametric {PDEs}},
  journal = {SMAI Journal of Computational Mathematics},
  year = {2021},
  volume = {7},
  pages = {121--157},
  doi = {10.5802/smai-jcm.74}
}

@article{LanthalerMishraKarniadakis2022,
  author = {Lanthaler, Samuel and Mishra, Siddhartha and Karniadakis, George E.},
  title = {Error Estimates for {DeepONets}: A Deep Learning Framework in Infinite Dimensions},
  journal = {Transactions of Mathematics and Its Applications},
  year = {2022},
  volume = {6},
  number = {1},
  pages = {tnac001},
  doi = {10.1093/imatrm/tnac001}
}

@article{Lanthaler2023PCANet,
  author = {Lanthaler, Samuel},
  title = {Operator Learning with {PCA-Net}: Upper and Lower Complexity Bounds},
  journal = {Journal of Machine Learning Research},
  year = {2023},
  volume = {24},
  number = {318},
  pages = {1--67},
  url = {https://jmlr.org/papers/v24/23-0478.html}
}

@inproceedings{GuptaEtAl2021Multiwavelet,
  author = {Gupta, Gaurav and Xiao, Xiongye and Bogdan, Paul},
  title = {Multiwavelet-based Operator Learning for Differential Equations},
  booktitle = {Advances in Neural Information Processing Systems},
  volume = {34},
  year = {2021},
  url = {https://papers.neurips.cc/paper_files/paper/2021/hash/c9e5c2b59d98488fe1070e744041ea0e-Abstract.html}
}

@article{TripuraChakraborty2023WNO,
  author = {Tripura, Tapas and Chakraborty, Souvik},
  title = {Wavelet Neural Operator for Solving Parametric Partial Differential Equations in Computational Mechanics Problems},
  journal = {Computer Methods in Applied Mechanics and Engineering},
  volume = {404},
  eid = {115783},
  year = {2023},
  doi = {10.1016/j.cma.2022.115783}
}

@article{FanaskovOseledets2022SNO,
  author = {Fanaskov, Vladimir and Oseledets, Ivan},
  title = {Spectral Neural Operators},
  journal = {Doklady Mathematics},
  volume = {108},
  number = {Suppl. 2},
  pages = {S226--S232},
  year = {2023},
  doi = {10.1134/S1064562423701107}
}

@article{Namias1980,
  author = {Namias, Victor},
  title = {The Fractional Order {Fourier} Transform and Its Application to Quantum Mechanics},
  journal = {IMA Journal of Applied Mathematics},
  year = {1980},
  volume = {25},
  number = {3},
  pages = {241--265},
  doi = {10.1093/imamat/25.3.241}
}

@article{TiwariEtAl2025CoNO,
  author = {Tiwari, Karn and Krishnan, N. M. Anoop and {Prathosh A. P.}},
  title = {{CoNO}: Complex Neural Operator for Continous Dynamical Physical Systems},
  journal = {APL Machine Learning},
  volume = {3},
  number = {2},
  eid = {026101},
  year = {2025},
  doi = {10.1063/5.0254013}
}

@article{WangEtAl2024FractionalFourierSeries,
  author = {Wang, Chen and Hou, Xianming and Wu, Qingyan and Dang, Pei and Fu, Zunwei},
  title = {Fractional {Fourier} Series on the Torus and Applications},
  journal = {Fractal and Fractional},
  year = {2024},
  volume = {8},
  number = {8},
  pages = {494},
  doi = {10.3390/fractalfract8080494}
}

@book{CorderoRodino2020,
  author = {Cordero, Elena and Rodino, Luigi},
  title = {Time-Frequency Analysis of Operators},
  series = {De Gruyter Studies in Mathematics},
  volume = {75},
  publisher = {De Gruyter},
  year = {2020},
  doi = {10.1515/9783110532456}
}

@article{AbdeljawadDittrich2026Frechet,
  author = {Abdeljawad, Ahmed and Dittrich, Thomas},
  title = {Approximation Rates in {Fr\'echet} Metrics: {Barron} Spaces, {Paley--Wiener} Spaces, and {Fourier} Multipliers},
  journal = {Neural Computation},
  volume = {38},
  number = {2},
  pages = {165--197},
  year = {2026},
  doi = {10.1162/NECO.a.1481}
}

@article{AbdeljawadCordero2026TimeFrequency,
  author = {Abdeljawad, Ahmed and Cordero, Elena},
  title = {Time-frequency analysis for neural networks},
  journal = {Applied and Computational Harmonic Analysis},
  volume = {84},
  pages = {101883},
  year = {2026},
  doi = {10.1016/j.acha.2026.101883}
}

@misc{AbdeljawadCarioniCordero2026,
  author = {Abdeljawad, Ahmed and Carioni, Marcello and Cordero, Elena},
  title = {Approximation Rates for Metaplectic Neural Networks},
  year = {2026},
  eprint = {2608.08872},
  archivePrefix = {arXiv}
}

@article{AbdeljawadGrohs2022Sobolev,
  author = {Abdeljawad, Ahmed and Grohs, Philipp},
  title = {Approximations with deep neural networks in {Sobolev} time-space},
  journal = {Analysis and Applications},
  volume = {20},
  number = {3},
  pages = {499--541},
  year = {2022},
  doi = {10.1142/S0219530522500014}
}

@misc{AbdeljawadDittrich2023SpaceTime,
  author = {Abdeljawad, Ahmed and Dittrich, Thomas},
  title = {Space-Time Approximation with Shallow Neural Networks in {Fourier Lebesgue} spaces},
  year = {2023},
  eprint = {2312.08461},
  archivePrefix = {arXiv}
}

@article{LanthalerStuart2026Complexity,
  author = {Lanthaler, Samuel and Stuart, Andrew M.},
  title = {The Parametric Complexity of Operator Learning},
  journal = {IMA Journal of Numerical Analysis},
  year = {2026},
  volume = {46},
  number = {2},
  pages = {647--712},
  doi = {10.1093/imanum/draf028}
}

@article{Hudzik06MatrixMultiplicationOperators,
  title = {Matrix Multiplication Operators on {{Banach}} Function Spaces},
  author = {Hudzik, H. and Kumar, Rajeev and Kumar, Romesh},
  date = {2006-02},
  journaltitle = {Proceedings of the Indian Academy of Sciences - Section A},
  shortjournal = {Proc. Indian Acad. Sci. (Math. Sci.)},
  volume = {116},
  number = {1},
  pages = {71--81},
  issn = {0370-0089},
  doi = {10.1007/BF02829740}
}

@article{LeshnoLinPinkusSchocken1993,
  author  = {Leshno, Moshe and Lin, Vladimir Ya. and Pinkus, Allan and Schocken, Shimon},
  title   = {Multilayer Feedforward Networks with a Nonpolynomial Activation Function Can Approximate Any Function},
  journal = {Neural Networks},
  volume  = {6},
  number  = {6},
  pages   = {861--867},
  year    = {1993},
  doi     = {10.1016/S0893-6080(05)80131-5}
}

@book{AdamsFournier2003,
  author    = {Robert A. Adams and John J. F. Fournier},
  title     = {Sobolev Spaces},
  edition   = {2},
  series    = {Pure and Applied Mathematics},
  volume    = {140},
  publisher = {Elsevier/Academic Press},
  address   = {Amsterdam},
  year      = {2003},
  isbn      = {978-0-12-044143-3}
}

@article{Li1996,
	author  = {Li, Xin},
	title   = {Simultaneous Approximations of Multivariate Functions and Their Derivatives by Neural Networks with One Hidden Layer},
	journal = {Neurocomputing},
	volume  = {12},
	number  = {4},
	pages   = {327--343},
	year    = {1996},
	doi     = {10.1016/0925-2312(95)00070-4}
}

@article{CorderoNicolaRodino2015,
  author  = {Elena Cordero and Fabio Nicola and Luigi Rodino},
  title   = {Integral Representations for the Class of Generalized Metaplectic Operators},
  journal = {Journal of Fourier Analysis and Applications},
  volume  = {21},
  number  = {4},
  pages   = {694--714},
  year    = {2015},
  doi     = {10.1007/s00041-014-9384-8}
}

@book{deGosson2011,
  author    = {Maurice A. de Gosson},
  title     = {Symplectic Methods in Harmonic Analysis and in Mathematical Physics},
  series    = {Pseudo-Differential Operators: Theory and Applications},
  volume    = {7},
  publisher = {Birkh{\"a}user},
  address   = {Basel},
  year      = {2011},
  doi       = {10.1007/978-3-7643-9992-4},
  isbn      = {978-3-7643-9991-7}
}

@book{Folland1989,
  author    = {Gerald B. Folland},
  title     = {Harmonic Analysis in Phase Space},
  series    = {Annals of Mathematics Studies},
  volume    = {122},
  publisher = {Princeton University Press},
  address   = {Princeton, NJ},
  year      = {1989},
  isbn      = {978-0-691-08528-9}
}

@article{HornikStinchcombeWhite1990,
  author  = {Kurt Hornik and Maxwell Stinchcombe and Halbert White},
  title   = {Universal Approximation of an Unknown Mapping and Its Derivatives Using Multilayer Feedforward Networks},
  journal = {Neural Networks},
  volume  = {3},
  number  = {5},
  pages   = {551--560},
  year    = {1990},
  doi     = {10.1016/0893-6080(90)90005-6}
}

@article{KovachkiLanthalerMishra2021,
  author  = {Nikola Kovachki and Samuel Lanthaler and Siddhartha Mishra},
  title   = {On Universal Approximation and Error Bounds for Fourier Neural Operators},
  journal = {Journal of Machine Learning Research},
  volume  = {22},
  number  = {290},
  pages   = {1--76},
  year    = {2021},
  url     = {https://www.jmlr.org/papers/v22/21-0806.html}
}

@article{LanthalerLiStuart2025,
  author  = {Samuel Lanthaler and Zongyi Li and Andrew M. Stuart},
  title   = {Nonlocality and Nonlinearity Implies Universality in Operator Learning},
  journal = {Constructive Approximation},
  volume  = {62},
  pages   = {261--303},
  year    = {2025},
  doi     = {10.1007/s00365-025-09718-3}
}

@inproceedings{LiEtAl2021FNO,
  author    = {Zongyi Li and Nikola Kovachki and Kamyar Azizzadenesheli and Burigede Liu and Kaushik Bhattacharya and Andrew M. Stuart and Anima Anandkumar},
  title     = {Fourier Neural Operator for Parametric Partial Differential Equations},
  booktitle = {International Conference on Learning Representations},
  year      = {2021},
  url       = {https://openreview.net/forum?id=c8P9NQVtmnO}
}

@book{RunstSickel1996,
  author    = {Thomas Runst and Winfried Sickel},
  title     = {Sobolev Spaces of Fractional Order, Nemytskij Operators, and Nonlinear Partial Differential Equations},
  series    = {De Gruyter Series in Nonlinear Analysis and Applications},
  volume    = {3},
  publisher = {Walter de Gruyter},
  address   = {Berlin},
  year      = {1996},
  doi       = {10.1515/9783110812411},
  isbn      = {978-3-11-015113-8}
}

@article{DeRyckLanthalerMishra2021Tanh,
  author = {De Ryck, Tim and Lanthaler, Samuel and Mishra, Siddhartha},
  title = {On the Approximation of Functions by {tanh} Neural Networks},
  journal = {Neural Networks},
  volume = {143},
  pages = {732--750},
  year = {2021},
  doi = {10.1016/j.neunet.2021.08.015}
}

@article{Barron1993,
  author  = {Barron, Andrew R.},
  title   = {Universal Approximation Bounds for Superpositions of a Sigmoidal Function},
  journal = {IEEE Transactions on Information Theory},
  volume  = {39},
  number  = {3},
  pages   = {930--945},
  year    = {1993},
  doi     = {10.1109/18.256500}
}

@article{Candes1999Harmonic,
  author  = {Cand{\`e}s, Emmanuel J.},
  title   = {Harmonic Analysis of Neural Networks},
  journal = {Applied and Computational Harmonic Analysis},
  volume  = {6},
  number  = {2},
  pages   = {197--218},
  year    = {1999},
  doi     = {10.1006/acha.1998.0248}
}

@article{SiegelXu2020GeneralActivation,
  author  = {Siegel, Jonathan W. and Xu, Jinchao},
  title   = {Approximation Rates for Neural Networks with General Activation Functions},
  journal = {Neural Networks},
  volume  = {128},
  pages   = {313--321},
  year    = {2020},
  doi     = {10.1016/j.neunet.2020.05.019}
}
\end{document}